\documentclass[sn-apa]{sn-jnl}

\usepackage{graphicx}%
\usepackage{multirow}%
\usepackage{amsmath,amssymb,amsfonts}%
\usepackage{amsthm}%
\usepackage{mathrsfs}%
\usepackage[title]{appendix}%
\usepackage{xcolor}%
\usepackage{textcomp}%
\usepackage{manyfoot}%
\usepackage{booktabs}%
\usepackage{algorithm}%
\usepackage{algorithmicx}%
\usepackage{algpseudocode}%
\usepackage{listings}%

\usepackage[T1]{fontenc}
\usepackage{lmodern}
\usepackage{textcomp}
\usepackage{anyfontsize}

\usepackage{wrapfig}
\usepackage[framemethod=TikZ]{mdframed}
\usepackage{lipsum}
\definecolor{myblue}{HTML}{4E7AC7}
\usepackage{pifont}
\usepackage{multicol}
\usepackage{svg}

\newcommand{\defeq}{\mathrel{:\mkern-0.25mu=}}

\begin{document}

\title[Learning to Optimize Feedback]{Learning to Optimize Feedback for One Million Students: Insights from Multi-Armed and Contextual Bandits in Large-Scale Online Tutoring}

\author*[1]{\fnm{Robin} \sur{Schmucker}}\email{rschmuck@cs.cmu.edu}

\author[2]{\fnm{Nimish} \sur{Pachapurkar}}

\author[2]{\fnm{Shanmuga} \sur{Bala}}

\author[2]{\fnm{Miral} \sur{Shah}}\email{miral.shah@ck12.org}

\author[1]{\fnm{Tom} \sur{Mitchell}}\email{tom.mitchell@cs.cmu.edu}

\affil[1]{
\orgname{Carnegie Mellon University},
\orgaddress{
\city{Pittsburgh},
\state{PA},
\country{USA}}}

\affil[2]{
\orgname{CK-12 Foundation},
\orgaddress{
\city{Menlo Park},
\state{CA},
\country{USA}}}

\abstract{
We present an online tutoring system that learns to provide effective feedback to students after they answer questions incorrectly. Using data from one million students, the system learns which assistance action (e.g., one of multiple hints) to provide for each question to optimize student learning. Employing the multi-armed bandit (MAB) framework and offline policy evaluation, we assess 43,000 assistance actions, and identify trade-offs between assistance policies optimized for different student outcomes (e.g., response correctness, session completion). We design an algorithm that for each question decides on a suitable policy training objective to enhance students' immediate second attempt success and overall practice session performance. We evaluate the resulting MAB policies in 166,000 practice sessions, verifying significant improvements in student outcomes. While MAB policies optimize feedback for the overall student population, we further investigate whether contextual bandit (CB) policies can enhance outcomes by personalizing feedback based on individual student features (e.g., ability estimates, response times). Using causal inference, we examine (i) how effects of assistance actions vary across students and (ii) whether CB policies, which leverage such effect heterogeneity, outperform MAB policies. While our analysis reveals that some actions for some questions exhibit effect heterogeneity, effect sizes may often be too small for CB policies to provide significant improvements beyond what well-optimized MAB policies that deliver the same action to all students already achieve. We discuss insights gained from deploying data-driven systems at scale and implications for future refinements. Today, the teaching policies optimized by our system support thousands of students daily.
}



\maketitle


\section{Introduction}
\label{sec:introduction}

The process of building intelligent tutoring systems (ITSs) involves many design decisions, ranging in granularity from specifying general instructional principles (e.g., interleaved vs. focused practice) to curating lesson materials, to writing practice questions and hints~\citep{Koedinger2013:Instructional}. Domain experts consider multiple design choices (e.g., different hints for a given question) but often find it difficult to predict which choice is best for students~\citep{Nathan2001:Expert}. In this context, data-driven design leverages system usage data to evaluate how different choices within the ITS affect students' learning processes and continuously improve the system over time~\citep{Koedinger2013:New}.

The ITS workflow is commonly structured into an outer loop, which focuses on activity sequencing, and an inner loop, which focuses on feedback and support~\citep{Vanlehn2006:Behavior}. Within these loops, ITSs employ teaching policies to make instructional decisions for individual students. Mastery Learning is a prominent example of an outer-loop policy that uses knowledge tracing to estimate students' proficiency and employs expert rules to sequence activities, developing proficiency one skill at a time~\citep{Ritter2016:How}. ITSs can also learn effective teaching policies directly from student data. For example, multi-armed bandit (MAB) algorithms leverage system usage data to identify instructional decisions that yield optimal learning outcomes for the overall student population. Beyond optimizing for the overall population, contextual bandit (CB) algorithms personalize instructional decisions based on individual student attributes (e.g., prior knowledge). Reinforcement learning further extends this personalization by identifying potential synergies between multiple instructional decisions distributed across the teaching process~\citep{Doroudi2019:Where}.

In this paper we describe an online tutoring system that employs a data-driven approach to optimize feedback policies providing assistance (e.g., hints and explanatory paragraphs) to students after they answer practice questions incorrectly. We report findings from a large-scale study analyzing data from one million students, evaluating the impact of individual assistance actions and assistance policies on various learning outcomes (e.g., reattempt correctness, session completion). This paper extends our initial work~\citep{Schmucker23:Learning} by exploring insights gained from optimizing and deploying MAB policies across six online science courses. Furthermore, it introduces causal machine learning methodologies to assess whether CB policies that personalize feedback based on individual student attributes can improve learning outcomes compared to MAB policies, which select feedback optimal for the overall student population. We discuss our methodology, system design, and various insights for the development of future learning technologies. The main contributions of this work include:

\begin{figure*}[t]
    \centering
    \includegraphics[width=\textwidth]{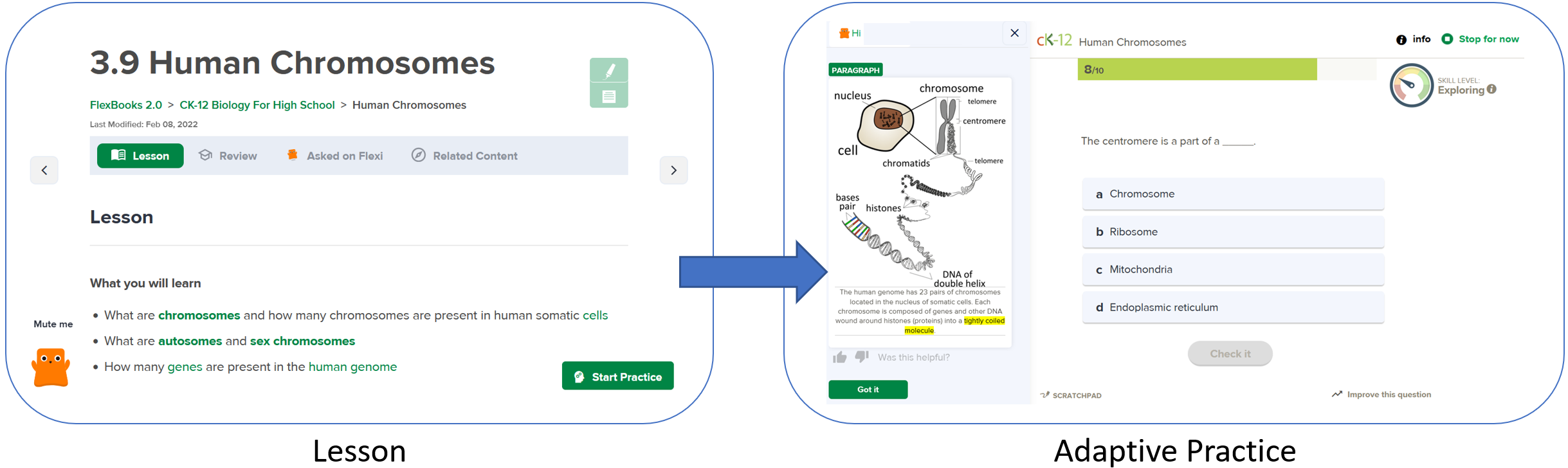}
    \vspace{-4mm}
    \caption[Overview of CK-12 platform and Adaptive Practice system.]{Example views from the biology concept \textit{Human Chromosomes}. [Left] In the \textit{Lesson} section the student interacts with multi-modal learning materials. [Right] During \textit{Adaptive Practice} the student develops and tests their understanding by answering practice questions. In the shown example the system displays a paragraph with illustration to assist the student after an initial incorrect response before the student reattempts the question.}
    \label{fig:ck12}
\end{figure*}

\begin{itemize}
    \item \textit{Quantifying effects of assistance.} We evaluate effects of over $43,000$ assistance actions on a variety of learning outcome measures (e.g., practice completion) using data from one million students. We study the relationships among different outcome measures and design a policy training algorithm that for each question decides on the most suitable training objective to optimize student success at the current question as well as overall session performance.
    \item \textit{Offline policy optimization.} We compute statistically significant estimates on the effects of multi-armed bandit policies trained to optimize different learning outcome measures. Studying assistance actions selected by these policies, we find that there is no single best assistance type (e.g., hint, keyword definitions clarifying the question text). We further find that assistance actions that benefit students the most when reattempting the current question are not always best for promoting performance on future questions.
    \item \textit{Large-scale A/B evaluation.} We evaluate the assistance policy trained using our algorithm in comparison to a randomized assistance policy in live use in over $166,000$ learning sessions. We find that the system's ability to learn to teach better using data from prior students improves learning outcomes of future students significantly.
    \item \textit{Assessing effect heterogeneity of assistance}: By combining causal inference with large-scale data from randomized experiments, we evaluate the extent to which the effects of individual assistance actions vary across students (heterogeneous treatment effect (HTE)). Considering various outcome measures, we assess the prevalence of HTEs with respect to student covariates, ranging from ability estimates to response times to overall ITS usage.
    \item \textit{Assessing potential of contextual policies}: We evaluate how contextual bandit (CB) policies, which consider individual student covariates to personalize decision-making, can leverage HTEs to improve outcomes over multi-armed bandit (MAB) policies. Using the potential outcomes framework, we examine the impact of CB policies that decide whether to provide assistance actions differing from the ones optimal for the overall student population.
\end{itemize}

\section{Related Work}
\label{sec:related_work}

We discuss related works on quantifying the effects of instructional actions and design choices within intelligent tutoring systems (ITSs) based on interaction log data. We also examine connections to ongoing efforts towards optimizing instructional policies using bandit and reinforcement learning algorithms and their relationships to heterogeneous treatment effect estimation.

\subsection{Evaluating Treatment Effects inside ITSs}

Initially the effects of ITSs on student learning were evaluated at the \textit{system level} by comparing a group of students that uses an ITS to a control group based on post-test results \citep{Kulik2016:Effectiveness}. Later research focuses on assessing the effects of individual \textit{instructional design choices} \citep{Koedinger2013:Instructional} within ITSs by studying students interacting with different configurations of the same tutoring system (e.g., \cite{Vanlehn2011:Relative,Selent2016:Assistments,Mclaren2022:Instructional,Nagashima2022:Does}). With the increasing adoption of online tutoring systems, large-scale log data has become available, which enables investigating the effects of more detailed system design choices, up to the choice of individual \textit{practice questions} and \textit{hints}.

As part of this development, the ASSISTments ecosystem \citep{Heffernan2014:Assistments} introduced AXIS \citep{Williams2016:AXIS}, the E-TRIALS TestBed \citep{Ostrow2017:Tomorrow} and the TeacherASSSIST system \citep{Patikorn2020:Effectiveness} to allow educators and researchers to create and evaluate the effectiveness of different problem sets and on-demand assistance materials. In the context of massive open online courses (MOOCs), DynamicProblem \citep{Williams2018:Enhancing} was introduced as a proof-of-concept system that supports the deployment of bandit algorithms to collect feedback from students by asking them about the helpfulness of individual support materials. Relatedly, the MOOClet framework \citep{Reza2021:MOOclet} allows instructors to specify multiple versions of educational resources and to evaluate them in A/B tests using randomization and bandit algorithms. Carnegie Learning introduced the UpGrade system \citep{Fancsali2022:Closing} as a flexible A/B testing framework designed to allow easy integration into various learning platforms. In a parallel development, EASI (Experiment as a Service)~\citep{Musabirov2024:Platform} was introduced as a cross-platform software facilitating the integration and deployment of adaptive experiments. In a series of deployments within the Open Learning Initiative (OLI) Torus \citep{Bier2023:Oli} platform, EASI interacted with $2,295$ students and evaluated different interventions to improve student engagement and participation in practice activities.

Relatedly, this paper describes a fielded online tutoring system at \text{CK12.org}, that learns to provide effective feedback (e.g., choose one of multiple available hints) to support students after they answer a practice question incorrectly, but before they reattempt the question. We use offline policy evaluation techniques \citep{Li2011:Unbiased} to leverage log data from over $23,800,000$ assistance requests from one million students in six science courses. The unprecedented scale of this data enables us to produce statistically significant and unbiased estimates of the effects of \textit{individual} assistance actions on different measures of student learning outcomes. Using insights from these analyses, we design a reward function and train multi-armed bandit policies \citep{Lattimore2020:Bandit} that optimize the student's success at answering the current question as well as their overall session performance. We further evaluate the effectiveness of the learned bandit policy in live use in over $166,000$ practice sessions, showing significant improvements in the system's ability to provide students with effective feedback during practice activities.

The adoption of learning technologies raises the question of whether all students benefit equally or if the benefits vary among individuals, a phenomenon known as heterogeneous treatment effect (HTE) \citep{Schmid2014:Effects,Schudde2018:Heterogeneous}. Understanding this variability is essential to ensure fairness and equity, as algorithmic and technological advancements should reduce existing disparities rather than exacerbate them \citep{Kizilcec2022:Algorithmic,Baker2022:Algorithmic}. Understanding the diverse effects of pedagogical decisions is further critical for personalized instruction, as it enables adaptive teaching policies that ensure that each student receives the support that benefits them most \citep{Bernacki2021:Systematic}.

In the ITS context, efficacy studies of the Cognitive Tutor Algebra I system highlighted how effects on learning outcomes can vary across individuals and linked the observed HTEs to differences in students' system usage patterns and demographics \citep{Ritter2013:Predicting,Pane2014:Effectiveness,Sales2016:Student}. More recent research employed causal inference methodologies and examined how students' problem-solving strategies modulate learning gains \citep{Leon2024:Estimating,Sales2024:Problem}. Other studies explored varying effects of specific instructional design choices, such as gamification, immediate versus delayed feedback and lecture video design, for students with differing academic and personal attributes \citep{Vanacore2023:Benefit,Prihar2023:Bandit,Vanacore2024:Effect,Pham2024:lool}. Focusing on fine-grained decisions in the ITS authoring process, the present paper combines causal machine learning with large-scale   log data to evaluate the extent to which the effects of \textit{individual} assistance actions vary across students.

\subsection{Data-Driven Feedback Policies}

Driven by the idea of optimizing instructional policies using student log data, bandit and reinforcement learning (RL) algorithms have found application in various educational contexts. Here, we provide a concise overview of research that uses data-driven algorithms to support students in the problem solving process (i.e., the ITS's inner loop). For a comprehensive review of RL for education we refer to surveys by \citet{Doroudi2019:Where} and \citet{Singla2021:Reinforcement}. 

\citet{Barnes2008:Toward} induced a Markov decision process (MDP) based on hundreds of student solution paths and used RL to generate new hints inside a logic ITS. Chi et al. \citep{Chi2009:To,Chi2010:Do} modeled a physics tutor via an MDP with $16$ states and learned a RL policy to improve student learning outcomes in a classroom setting by deciding whether to ask the student to reflect on a problem or to provide them with additional information. \citet{Georgila2019:Using} used Least-Squares Policy Iteration \citep{Lagoudakis2003Least} to learn a feedback policy for an interpersonal skill training system using data describing over 500 features from $72$ participants. \citet{Ju2020:Modeling} identified critical pedagogical decisions based on Q-value and reward function estimates derived from logs of $1,148$ students inside a probability ITS. Relatedly, Ausin et al. \citep{Ausin2019:Leveraging,Ausin2021:Tackling} explored Gaussian Process- and inverse RL-based approaches to address the credit assignment problem inside a logic ITS. A series of works \citep{Spain2019:Enhancing,Fahid2021:Adaptively,Spain2021:Decision} used a random policy to collect data from $500$ students in an operational command course and explored offline RL techniques to learn adaptive scaffolding policies based on the ICAP \citep{Chi2014:ICAP} framework. 

In contrast to the above works which largely focus on \textit{personalized} assistance action selection, this paper employs a multi-armed bandit approach that for each of $10,210$ questions learns to select the teaching action that is most effective \textit{on average} across all students, given that their first answer to this particular question was incorrect. We use offline policy evaluation techniques \citep{Li2011:Unbiased} to quantify the impact of each of $43,355$ assistance actions on different learning outcome measures (e.g., practice completion) and study relationships among individual measures.

Closely related is a recent study by \citet{Prihar2022:Automatic} that compared a multi-armed bandit algorithm based on Thompson Sampling to a randomized assistance policy with respect to their ability to increase students' \textit{success on the next question} by choosing effective support materials. The policies were trained to select from a content pool featuring hints and explanations some of which included images and videos. In their experiment with $2,923$ questions they found the multi-armed bandit algorithm to be only slightly more effective than the random policy and argued that this is due to sample size limitations (on average $6.5$ samples per assistance action). In contrast, our work is able to accurately estimate the impact of individual assistance actions on \textit{different measures of learning outcomes} by having access to hundreds of samples for individual actions (Figure \ref{fig:error_distribution}). Further, in contrast to \citet{Prihar2022:Automatic} we quantify treatment effects by automatically displaying assistance in response to incorrect student responses (Figure \ref{fig:question_workflow}) which avoids self-selection effects when assistance is only provided upon student request.

While RL policies are typically trained to optimize outcomes for the overall student population, recent research has explored how the effects of these policies can vary across individuals. \citet{Ausin2019:Leveraging} observed varying effects among high and low competency learners when training policies that select between worked examples and problem solving activities in a logic tutor. \citet{Abdelshiheed2023:Leveraging} optimized instructional policies to promote the adoption of meta-cognitive strategies and observed effect differences based on students' prior familiarity with these strategies. \citet{Leite2022:Heterogeneity} employed the causal random forest (CRF) algorithm \citep{Wager2018:Estimation} to study video recommendation policies within an algebra course unveiling heterogeneous treatment effects related to prior knowledge and socioeconomic factors. In the context of a narrative-based math learning environment, \citet{Nie2023:Understanding} proposed a methodology for analyzing decisions and effects of an RL policy across student subgroups with varying pre-test scores and math-anxiety levels. Within the same ITS, \citet{Ruan2024:Reinforcement} found that students with lower pre-test scores benefited the most from the RL optimized teaching policy.

Unlike the above, this paper focuses on understanding the extent to which the effects of individual assistance actions designed to support students after an incorrect response to a practice questions vary across individuals. We further assess the degree to which contextual policies can leverage these variations to improve learning outcomes over non-contextual policies. Prior studies comparing the effects of contextual and non-contextual bandit policies on learning outcomes include \cite{Li2020:Getting} and \cite{Prihar2022:Automatic}, though those were not centered on effect heterogeneity in problem-solving assistance. While these works suggest that personalization can be beneficial in certain settings, both rely on simulation studies using resampling techniques that can increase the risk of false discovery and findings that might not be representative of the true prevalence of effect heterogeneity within assistance selection decisions.

\section{Flexbook 2.0 System}
\label{sec:ck12}

The CK-12 Foundation is a US-based non-profit company that provides millions of students worldwide with access to free educational resources aligned to state curriculum standards. CK-12 actively develops and hosts the Flexbook 2.0 system\footnote{\url{https://www.ck12.org}}, a web-based ITS that offers a variety of courses targeting different subjects and grade levels. Each individual course consists of a sequence of \textit{concepts}. Each individual concept has a \textit{Lesson} section which offers learning materials in the form of texts, illustrations, videos and interactive elements, as well as an \textit{Adaptive Practice} (AP) section that allows students to practice and test their understanding (Figure \ref{fig:ck12}).

\begin{figure}[t]
    \centering
    \includegraphics[width=0.45\textwidth]{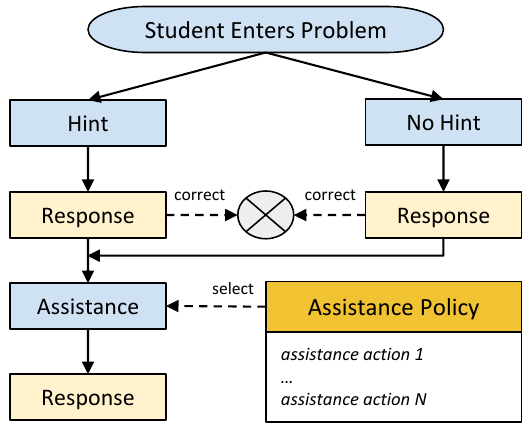}
    \caption{Problem solving workflow. The system learns an assistance policy that decides which of several candidate assistance actions (e.g., one of multiple hints) to provide to a student after they answer a practice question incorrectly before they reattempt the question.}
    \label{fig:question_workflow}
\end{figure}

From a high-level perspective, the AP section features an item response theory (IRT)-driven question sequencing system (outer loop) that tries to select practice questions that match the student's current ability level (\textit{Goldilocks} principle \citep{Koedinger2013:Instructional}). After the system selects a question, the student enters a problem solving workflow (inner loop) illustrated by Figure \ref{fig:question_workflow}. In the problem solving workflow the student has the option to request a hint before submitting a first response to the question. If the first response is incorrect, the system provides immediate feedback and support by displaying one \textit{assistance action} (e.g., a hint or important keyword definitions) and asks the student to reattempt the question. Afterwards, the system uses the student's first response to update the student's IRT-based ability estimate and selects the next question. This process repeats until the student completes the AP session successfully by achieving ten correct responses or until the question pool is exhausted in which case the student can reset and reattempt the AP session.

This paper centers on the question of how we can employ data-driven techniques to learn an \textit{assistance policy} that selects the most effective assistance action for each individual practice question (Figure \ref{fig:question_workflow}) with the goal of enhancing the AP system's ability to provide students with effective feedback. We consider six of CK-12's science courses, each used by hundreds of thousands of middle and high school students each year and whose content has been developed and refined for over ten years. The courses cover hundreds of concepts related to biology, chemistry, physics and earth science. Each practice question falls into one of five distinct categories:
\begin{itemize}
    \item \textit{Multiple-choice}: Questions with three or four options and a single correct answer.
    \item \textit{Select-all-that-apply}: Questions that require the student to select the correct subset of displayed response options.
    \item \textit{Fill-in-the-blank}: Questions that require the student to fill in missing words in a sentence or short paragraph.
    \item \textit{Short-answer}: Questions that require the student to write a free-form answer typically consisting of at most three words.
    \item \textit{True-false}: Multiple-choice questions with two options. Students cannot reattempt this type of question. The AP system selects non-true-false questions when available.
\end{itemize}

The AP system associates each practice question with a set of assistance actions that can be used to provide feedback to incorrect student responses. All assistance actions were created by human domain experts. Most non-true-false question are associated with 3 to 7 different assistance actions with some variations between the individual courses (Table \ref{tab:data_overview}). Individual teaching actions vary in information content and fall into one of six distinct categories: 
\begin{itemize}
    \item \textit{Hint}: A one or two sentence text or a relevant illustration designed to help the student reflect on the question.
    \item \textit{Paragraph}: A paragraph from the Lesson section that is relevant for the question. This can also contain a related illustration.
    \item \textit{Vocabulary}: Definitions of important keywords in the question text.
    \item \textit{Remove distractor}: Removes one possible option before asking the student to reattempt the question (only for multiple-choice and select-all-that-apply questions).
    \item \textit{First letter}: Displays first letter of the correct solution (only for fill-in-the-blank and short-answer questions).
    \item \textit{No assistance}: Prompts the student to reattempt the question without providing additional feedback. Serves as baseline for computing treatment effects of other assistance actions.
\end{itemize}

\section{Methodology}
\label{sec:methodology}

We start by formulating the task of learning to select effective assistance actions as a machine learning problem known as a multi-armed bandit problem. We then discuss the student log data collection process and describe the different measures of learning outcomes we consider. Finally, we present the offline reinforcement learning techniques we employ to optimize and evaluate multi-armed bandit and contextual bandit policies. We provide details on the software architecture underlying our system in Appendix \ref{app:system_architecture}. Descriptions of student context features are provided in Appendix \ref{app:features}.

\subsection{Formal Problem Statement}
\label{subsec:problem_statement}

We denote the set of practice questions inside the system as $Q = \{q_1, \dots, q_k\}$. Each question $q \in Q$ is associated with a set of $n_q$ assistance actions $A_q = \{a_{q,1}, \dots, a_{q,n_q}\}$ that the system can use to support students after their first incorrect response to $q$. Our system learns a distinct question-specific bandit policy $\pi_q$ for each distinct practice question $q \in Q$.  Each learned policy, $\pi_q$, outputs an assistance action $a_q \in A_q$.  When a student answers question $q$ incorrectly, this is the assistance action given to the student before they re-attempt answering question $q$.   The result for the student is captured by a real-valued reward signal $r_{a_q} \in \mathbb{R}$  indicating the effectiveness of that assistance action (e.g., whether the student answered question $q$ correctly on their second attempt, or the final score of the student on the current lesson).  These reward signals are used to train the policy $\pi_q$, and are assumed to be sampled from some initially unknown action-specific and time-invariant distribution $R_{a_q}$. The optimal \textit{multi-armed bandit} (MAB) policy $\pi_{q,MAB}^*$ maximizes the expected reward (expected students' learning outcomes) by selecting assistance action $a_{q, MAB}^* = \mbox{arg\,max}_{a_q \in A_q} \mathbb{E}[r_{a_q}]$.

While the MAB framework always selects the assistance action that is optimal \textit{on average} across all students, the \textit{contextual bandit} (CB) framework personalizes decision-making for student $s$ by considering their context features $x_s \in \mathbb{R}^n$.  Context features are designed to capture characteristics of the individual student (e.g., whether they correctly answered the previous question in this lesson).  The optimal CB policy $\pi_{q, CB}^*$ maximizes the expected outcome for student $s$ by selecting assistance action $a_{q, CB}^* = \mbox{arg\,max}_{a_q \in A_q} \mathbb{E}[r_{a_q} | x_s]$.  Note this CB policy takes the student-specific context features $x_s$ into account when selecting the assistance action, and CB policies can therefore improve outcomes over MAB policies.   However, this will only be the case if the best assistance action for question $q$ varies from one student to another (as discussed in Section \ref{subsec:cb_policy_eval}).

This bandit framework enables our system to automatically make design decisions by learning from the observed behavior of earlier students. It is difficult for human experts to predict the most effective design ahead of time \citep{Nathan2001:Expert}. Therefore, we adopted this bandit framework to learn the effects of potential design choices using student data, thereby refining the ITS design automatically over time. Section \ref{sec:discussion} discusses benefits and limitations of our bandit formulation.

\subsection{Data Collection}
\label{subsec:data_collection}


\begin{table}[!t]
\centering
\caption{Data collection overview. \textit{Shown questions} refers to the number of times a question was shown to a student. \textit{Student responses} refers to the total number of submitted responses, including both first and second attempts to answer the question. \textit{Shown actions} refers to the number of times an assistance action was shown to a student after an incorrect first response.  \textit{Average correctness} indicates the fraction of correct first responses. \textit{Average completion} indicates the fraction of sessions in which students achieved ten correct responses.}
\begin{tabular}{lrrrrrr}
\hline
& \texttt{Biology} & \texttt{Chemistry} & \texttt{Physics} & \texttt{Life Sci} & \texttt{Earth Sci} & \texttt{Phys Sci} \\
\hline
\# of questions & 12,496 & 7,538 & 1,780 & 4,833 & 4,797 & 5,018 \\
\# of assist. acts. & 36,354 & 24,980 & 6,190 & 17,464 & 17,505 & 16,714 \\
\# of concepts & 470 & 406 & 112 & 266 & 325 & 297 \\
\hline
\# of students & 191,554 & 143,958 & 52,284 & 191,370 & 212,324 & 203,686 \\
\# of sessions & 1,274,072 & 1,467,654 & 329,125 & 1,113,518 & 2,008,530 & 1,834,723 \\
\# of shown quests. & 15,582,835 & 16,247,815 & 3,420,751 & 12,703,418 & 21,608,431 & 19,563,538 \\
\# of responses & 20,425,691 & 21,049,821 & 4,367,063 & 16,713,933 & 26,399,756 & 23,994,522 \\
\# of shown acts. & 4,842,856 & 4,802,006 & 946,312 & 4,010,515 & 4,791,325 & 4,430,984 \\
\hline
avg. correctness & 63.3\% & 66.0\% & 68.5\% & 63.1\% & 68.4\% & 69.0\% \\
avg. completion & 75.1\% & 61.8\% & 53.3\% & 60.6\% & 49.1\% & 51.6\% \\
\hline
\# of eval. quests. & 1,336 & 2,355 & 551 & 1,815 & 2,107 & 2,046 \\
\# of eval. acts. & 7,707 & 8,451 & 1,786 & 7,351 & 9,821 & 8,239 \\
\# acts./question & 5.77 & 3.59 & 3.24 & 4.05 & 4.66 & 4.03 \\  
\hline
\end{tabular}
\label{tab:data_overview}
\end{table}

This study centers around six online science courses frequented by hundreds of thousands of middle and high school students each year. Because these courses have been in continuous refinement for over ten years, the course content bases contain \textit{multiple} assistance actions for individual questions. This raises the question of \textit{what type} of assistance is most effective (e.g., targeted hints or related keyword definitions?). On a more fine-granular basis, even if the domain experts decide to configure the system with a \textit{specific type} of assistance, it is still open \textit{which action} in the candidate pool is most effective (e.g., which exact hint among the available ones?).

\begin{figure}[t]
    \centering
    \includegraphics[width=0.495\columnwidth]{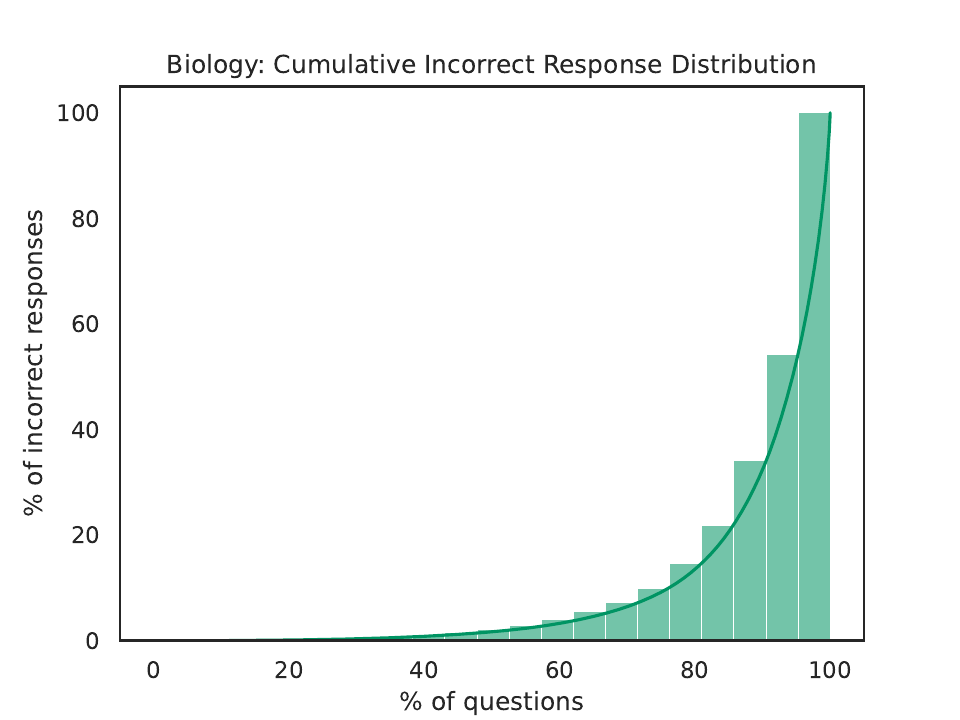}
    \hfill
    \includegraphics[width=0.495\columnwidth]{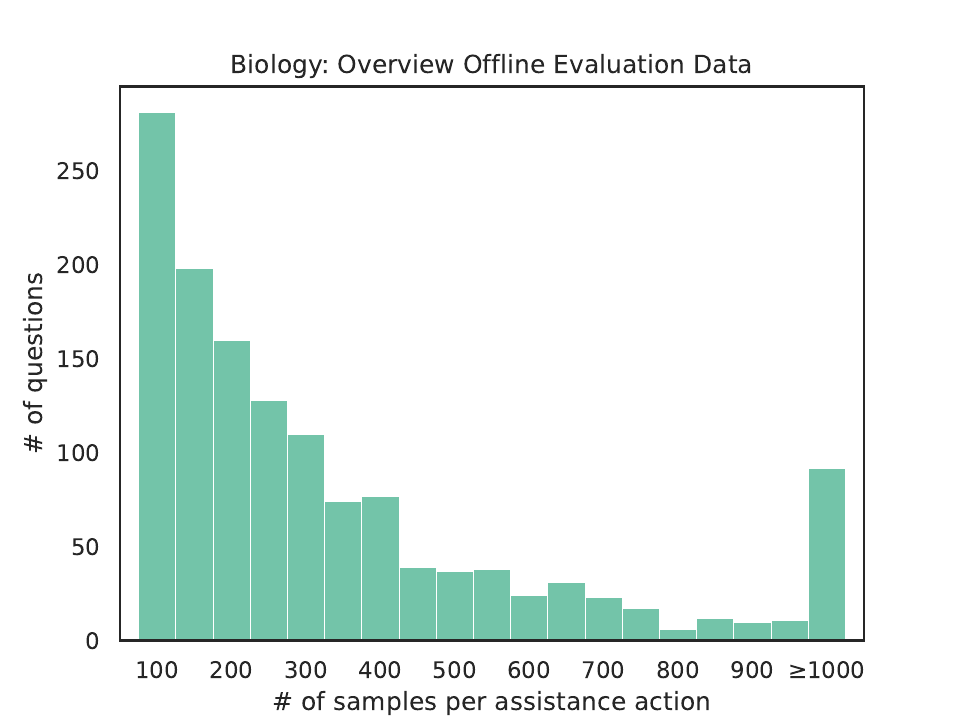}
    \caption{[Left] Cumulative distribution showing how incorrect student responses are distributed over all non-true-false questions ($n=7,599$) in the biology course. The $x$-axis indicates the proportion of questions sorted in ascending order based on number of incorrect responses. The $y$-axis indicates the cumulative proportion of incorrect responses for a subset of questions. We observe that $80\%$ of incorrect responses occur on only $15.4\%$ of questions. [Right] Histogram showing the number of samples available per assistance action for each of the $1,336$ biology questions studied in the offline evaluation experiments.}
    \label{fig:error_distribution}
\end{figure}

Responding to these questions, we conduct a large-scale evaluation to assess quantitatively the impact of \textit{individual} assistance actions on different measures of student learning outcomes. In a first step, from August 23rd, 2022 to January 11th, 2023, a randomized assistance policy was introduced into the problem solving workflow of the biology course (Figure \ref{fig:question_workflow}). Each time this policy is queried to provide assistance for a question $q \in Q$, it uniformly (with same chance) chooses one action at random from the set $A_q$. In the second step, after confirming the real world benefits of our methodology, we applied the same methodology to evaluate assistance actions inside the other five science courses from January 19th, 2023 to May 11th, 2023. An overview of the data collected for each individual course, is provided by Table \ref{tab:data_overview}. Overall, we collected logs from about one million students, that interacted with over $36,000$ questions and more than $119,000$ assistance actions. Overall, the randomized policy responded to over $23,800,000$ assistance queries. One interesting observation we made during the data collection process is that the majority of observed student errors occur on a rather small number of questions. For example, in the biology course $80\%$ of incorrect responses occur on only $15.4\%$ of questions (Figure \ref{fig:error_distribution}). This shows that we can respond to the majority of assistance queries that occur during student practice by learning policies for a smaller number of questions.

\subsubsection{Data Preprocessing}

Before assistance policy optimization and evaluation we perform the following preprocessing steps: (i) To avoid early using data from students who dropped out early, we only consider practice sessions in which students responded to at least five different questions. (ii) To avoid memorization effects, we only consider each student's first practice attempt for each concept. (iii) To avoid confounding, we estimate the effects of individual assistance actions using only practice sessions in which the student did not request a hint before their first attempt on that question. (iv) To achieve high confidence in our effect estimates we focus on practice questions with at least $100$ samples per assistance action. As a result, our evaluation of MAB policies considers a set of $10,210$ unique questions associated with $43,355$ actions. Our evaluation of CB policies focused on questions for which the \textit{no assistance} baseline action was included into the action set and considered a set of $1,753$ unique questions with $7,458$ actions.

\subsection{Measures of Learning Outcomes}
\label{subsec:outcome_measures}

A central decision in our data-driven design process is the definition of a reward function that takes as input log data generated by a student practice session and that outputs a reward value that quantifies the degree to which the assistance provided by the system promoted successful learning. The reward function guides the assistance policy training process by formally defining which learning outcome measures to focus on and shapes the experience of future students.

In conversations the designers of the practice system mentioned promoting growth in \textit{student knowledge} as well as \textit{student engagement} as their primary objectives. Unfortunately, student knowledge and engagement are both unobservable variables and the system is limited in that it can only access data that describes the student's observable interactions with the website interface. Because of this, we compiled--in close collaboration with domain experts--a list specifying different measures of learning outcomes that can be computed from observed student log data:

\begin{itemize}
    \item \textit{Reattempt correct}: Binary indicator ($\{0, 1\}$) of whether the student is correct when they reattempt the question after an assistance action.
    \item \textit{Student ability}:  A measure of student success for the entire lesson, based on their first responses to all questions they encounter.   We use the 3PL-IRT student ability estimate  \citep{Ayala2013:Theory}.
    \item \textit{Session success}: Binary indicator ($\{0, 1\}$) of whether the student achieves 10 correct responses in the practice session overall.
    \item \textit{Future correct rate}: Proportion of student's correct responses on first attempts on questions after an assistance action.
    \item \textit{Next quest. correct}: Binary indicator ($\{0, 1\}$) of whether the student is correct on the next question after the assistance action \citep{Prihar2022:Automatic}.
    \item\textit{Future response time}: Measures in seconds the student's average response time on questions after an assistance action (individual question response time values are capped at 60 seconds (95\% percentile) to mitigate outliers).
    \item \textit{Student confidence}: Three level indicator $(\{1, 2, 3\})$ of the student's self-reported confidence level at the end of the practice session.
\end{itemize}

We use student log data to evaluate the effects of individual assistance actions and policies on these different measures of learning outcomes. We further study the relationships between the individual measures to understand synergies and potential trade-offs. Based on insights from these analyses--discussed in detail in Section \ref{subsec:bandit_results}--we define our \textit{combined reward} function $R$ as
\begin{equation}
    R(s, q) = 0.4 \cdot \text{reattempt\_correct}(s, q) + 0.6 \cdot \text{student\_ability}(s).
    \label{eq:reward}
\end{equation}
Here, $s$ represents information collected during a student's entire practice session and $q$ indicates the question for which assistance was provided to the student. The reward value is computed as a weighted sum that considers the student's success at reattempting question $q$ as well as their overall performance in the practice session. The main motivation behind this function is that we want to provide assistance in a way that keeps the student engaged by aiding them in solving the current question, and that also conveys generalizable insights, that help the student solve other questions in the practice session.

\subsection{Evaluation of Multi-Armed Bandit Policies}
\label{subsec:offline_evaluation}

We present the offline reinforcement learning methodology used to estimate effects of individual assistance actions, train our multi-armed bandit policies and evaluate their effects on student learning with respect to different outcome measures.

\subsubsection{Policy Optimization}

To train and evaluate the effects of different assistance policies without conducting repeated live experiments we rely on offline policy optimization \citep{Li2011:Unbiased} and leverage the log data collected by the randomized exploration policy (Section \ref{subsec:data_collection}). First, we estimate the effectiveness of individual assistance actions by computing the mean value for each of the learning outcome measures across all relevant practice sessions. From there, our experiments study various multi-armed bandit policies trained to optimize different learning outcome measures. In preliminary experiments, we found that when using measures with high variance as training objectives (i.e., \textit{student ability} and \textit{session success}), the conventional policy optimization approach (that for each question selects the assistance action estimated to be optimal) struggles to reliably identify actions that perform well in evaluations on separate test data. For the average question we found optimizing policies for \textit{reattempt correctness} (a measure with focus on a single question and thus lower variance) to be the most effective way to also boost \textit{student ability} and \textit{session success} due to its positive correlations to the other measures (Figure \ref{fig:signal_correlations} left).

Still, for a sizeable number of questions the conventional approach yielded better policies when directly optimizing for our \textit{combined reward} function (Equation~\ref{eq:reward}). These tended to be questions with more available data or with larger differences in the effects of individual assistance actions. This motivated the design of a training algorithm that for each question decides whether we have sufficient data to optimize the \textit{combined reward} directly or whether we should use the low variance \textit{reattempt correctness} measure. We first use the training data to identify the two actions that optimize \textit{combined reward} and \textit{reattempt correctness} respectively. We then conduct a one-sided Welch $t$-Test to decide whether the former has a significantly larger effect on the \textit{combined reward} than the \textit{reattempt correctness} action and if not select the low variance \textit{reattempt correctness} measure as the question-specific training objective.

\subsubsection{Policy Evaluation}

In the offline evaluation experiments we report mean performance estimates derived from a $20$ times repeated $5$-fold cross validation (Section \ref{subsec:bandit_results}). In each fold $80\%$ of practice sessions are used for policy training and the remaining $20\%$ are used for testing. This process yields a statistically unbiased estimate of the bandit policy's performance as it simulates a series of interactions with different students inside the system and avoids overfitting effects of sampling with replacement-based approaches \citep{Li2011:Unbiased}. For the significance test we determine a suitable $p$-value for each individual outcome measure by evaluating $p \in \{0.01, 0.02, \dots, 0.1\}$ via cross-validation. The final policy used in live A/B evaluation is trained using data from all practice sessions and optimizes our reward function as defined by Equation \ref{eq:reward}.

\subsection{Evaluation of Contextual Bandit Polices}
\label{subsec:contextual_methodology}

Here we present the causal machine learning methodology we used to assess contextual bandit policies. First, we introduce the potential outcomes framework \citep{Rubin2005:Causal} describing relationships between learning personalized assistance policies and treatment effect estimation. From there, we discuss hypothesis tests for the detection of linear and non-linear treatment effect heterogeneity in our data. Lastly, we describe a procedure for assessing the degree to which contextual assistance policies can leverage effect heterogeneity to improve outcomes over non-contextual policies. Descriptions of all considered student context features are provided in Appendix \ref{app:features}.

\subsubsection{Potential Outcomes Framework}

We seek to understand how the decisions we make about providing a particular assistance action affects the individual student. To study this question we employ the potential outcomes framework \citep{Rubin2005:Causal}. Formally, we model each sample in our dataset as a triple $(X_i, W_i, Y_i)$. Here $X_i \in \mathbb{R}^d$ is the covariate vector for student $i$, $W_i \in \{0, 1\}$ is an indicator describing whether $i$ received treatment or control and $Y_i \in \mathbb{R}$ is the learning outcome (e.g., session success or student ability). For us $W_i = 1$ marks students receiving the assistance action and $W_i = 0$ marks students receiving a control intervention (e.g., ``no assistance''). Estimating causal effects is challenging because each student is only observed in one treatment state, i.e., they either received the treatment or the control. To reason about causal effects, the potential outcomes framework associates each individual with two random variables ($Y_i(1)$, $Y_i(0)$). Depending on the treatment state, the outcome for student $i$ is observed as
\begin{equation}
Y_i \defeq Y_i(W_i) = \begin{cases} 
Y_i(1) & \text{if } W_i = 1 \text{ (treated)} \\
Y_i(0) & \text{if } W_i = 0 \text{ (control)}.
\end{cases}
\end{equation}

\begin{figure}[t]
    \centering
    \includegraphics[width=\linewidth]{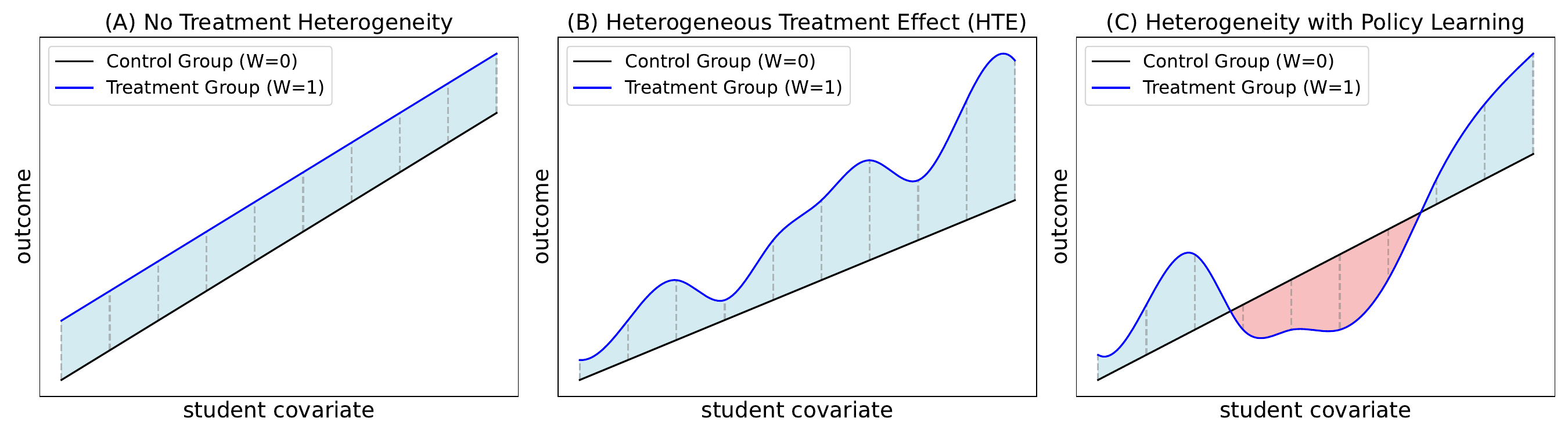}
    \caption{Illustration of heterogeneous treatment effect (HTE). (A) The treatment effect is constant for all students. (B) We observe HTE in that the effects of the treatment vary based on the student covariate. The conditional treatment effect is positive for \textit{all} students. (C) We observe HTE and the treatment is beneficial only for a \textit{subset} of students. The optimal policy maximizes learning outcomes by making treatment decisions informed by the student covariate.}
    \label{fig:conceptual}
\end{figure}

Because we never observe both $Y_i(1)$ and $Y_i(0)$ for the same student, we cannot estimate individual treatment effects $Y_i(1) - Y_i(0)$ directly. Instead, our analysis utilizes the fact that our assistance dataset was collected in a randomized experiment which ensures independence of the observed outcomes and the treatment assignments ($Y_i(1),\, Y_i(0) \perp W_i$). This enables us to estimate the average treatment effect (ATE), defined as
\begin{equation}
    \tau \defeq \mathbb{E} \left[ Y_i(1) - Y_i(0) \right],
    \label{eq:ate_tau}
\end{equation}
by computing the difference in mean outcomes between students in the treatment and the control group. Reflecting on Section \ref{subsec:offline_evaluation}, the multi-armed bandit (MAB) policies focused on evaluating a set of available assistance actions $A = \{a_{1}, \dots, a_{n}\}$ to learn a policy that selects action $a^* = \mbox{arg\,max}_{a \in A} \mathbb{E}[Y_{i,a}(1)]$ optimizing expected learning outcomes for the overall student population. This is equivalent to selecting the action with highest ATE $\tau_a$ in the potential outcomes framework (assuming equal controls). Here, we center on the question of whether what is best for the overall student population is also best for the individual student. Hence, we study heterogeneous treatment effects (HTEs) by estimating the conditional average treatment effect (CATE) function defined as 
\begin{equation}
    \tau(x) \defeq \mathbb{E} \left[ Y_i(1) - Y_i(0) \,|\, X_i = x \right].
\end{equation}
More specifically, CATE describes the treatment effect for each student by considering their personal covariate vector $x \in \mathbb{R}^d$. Importantly, estimating the CATE function enables us to address two central questions: (i) Does the effectiveness of individual assistance actions vary across the students population (do HTEs exist)? This is important because HTEs are a prerequisite for effective personalization (Figure \ref{fig:conceptual}). (ii) Can assistance policies that consider potential HTEs improve learning outcomes over assistance policies that only consider ATEs (MAB policies)? 

In the following we will approach these questions by studying two types of treatment decisions. First, we focus on assessing the existence of HTEs in our decisions by analyzing outcomes of students that received a particular assistance action (treatment) and students that received ``no assistance'' (control). Second, we gauge the degree to which we can improve outcomes by learning contextual bandit (CB) policies that deviate from the MAB policies (Section \ref{subsec:offline_evaluation}). Here, we study the decision whether a student should receive a particular assistance action (treatment) different from the action that is best for the overall student population (control).

\subsubsection{Evaluation of Heterogeneous Treatment Effect}

We introduce our methodology for detecting HTE in assistance decisions--a prerequisite for effective personalization. For each assistance action in our dataset, we study outcomes of students receiving that assistance action (treatment) and students receiving ``no asistance'' (control). We perform hypothesis tests related to linear and non-linear interactions between student covariates and learning outcomes via regression and non-parametric causal machine learning algorithms \citep{Wager2018:Estimation}. Because our study involves thousands of tests we control the false discovery rate (FDR) at $0.2$ via the BH adjustment \citep{Benjamini1995:Controlling}.

\textbf{Linear HTE}
We assess linear interactions between assistance outcomes and student covariates via regression analysis. For each individual assistance action and individual student covariate $x_i \in \mathbb{R}$ (e.g., ability value, response time) we fit a linear regression model $f$ predicting learning outcomes as 
\begin{equation}
    f(Y_i \,|\, X_i) = \beta_0 + \beta_w W_i + \beta_x x_i + \beta_{wx} W_i x_i.
    \label{eq:jedm24_linear_heterogeneity}
\end{equation}
Here parameter $\beta_0 \in \mathbb{R}$ models average outcome for students in the control condition, $\beta_w \in \mathbb{R}$ is average treatment effect (equivalent to $\tau$ in Equation~\ref{eq:ate_tau}), $\beta_x \in \mathbb{R}$ captures potential effects attributed to the covariate but not the treatment, and $\beta_{wx} \in \mathbb{R}$ captures effects attributed to interactions between the covariate and the treatment (i.e., effect heterogeneity). To test for HTE we fit the regression model to student data and assess whether $\beta_{wx}$ is significantly different from zero. Note, that while Equation \ref{eq:jedm24_linear_heterogeneity} describes a linear regression, we employ an analog logistic regression formulation to study binary outcome variables (i.e., reattempt correctness, session success).

\textbf{Non-Linear HTE}
To assess non-linear treatment effect heterogeneity, we employ the causal random forest (CRF) algorithm, proposed by \citet{Wager2018:Estimation}. CRF extends the traditional random forest method to provide unbiased CATE estimates and valid confidence intervals, enabling insights into treatment effects across a heterogeneous population. The key idea behind the algorithm is honest estimation of CATE values through sample splitting. The data is divided into two disjoint subsets--one for constructing the tree structure (the training sample) and another for estimating treatment effects within each leaf (the estimation sample). The CRF algorithm constructs an ensemble of such trees, each partitioning the student population into groups with similar covariates. By separating the data used for tree splitting (group identification) from the data used for estimation, the algorithm can produce unbiased CATE estimates. 
For each assistance action we fit a CRF to model learning outcomes of students who received that particular teaching action, compared to  students who received ``no assistance''. For this we employ the `grf' package \citep{Tibshirani2021:GRF} and optimize hyperparameters in random search over $100$ parameter configuration via cross-validation. Using the resulting CRF models we assess HTE in two ways:
\begin{itemize}
    \item \textit{Calibration Test}: We conduct a calibration test to determine whether the treatment effects estimated by the CRF model are predictive of observed outcomes beyond the average treatment effect (ATE) \citep{Cameron2015:Practitioner}. Specifically, we test the null hypothesis that the estimated treatment effects are uncorrelated with the residuals of the outcome model. This is done by regressing the residuals on both the mean CRF prediction and the individual HTEs. If the slope of this regression is significantly different from zero, it indicates that the estimated treatment effects capture real variations in assistance effects across the student population, thus providing evidence for treatment effect heterogeneity.
    \item  \textit{RATE}: The rank-weighted average treatment effect (RATE) method \citep{Yadlowsky2024:Evaluating} assesses a model's ability to rank individuals based on their predicted benefits from a treatment. To compute RATE, we first rank individuals based on their estimated CATEs from the CRF model. We then calculate a weighted average of the observed treatment effects, where the weights are a function of the individuals' ranks. Specifically, higher-ranked individuals (those with higher estimated CATEs) receive greater weight in the calculation. The RATE metric effectively measures the expected treatment effect among individuals who are most likely to benefit from the treatment according to the model. We test for non-linear effect heterogeneity by comparing the RATE obtained from the CRF model to that from a model that assumes uniform treatment effects. A significantly higher RATE from the CRF model indicates that it successfully identifies individuals with higher treatment effects, thus providing evidence for treatment effect heterogeneity in our assistance decisions.
\end{itemize}

\subsubsection{Evaluation of Contextual Assistance Policies}
\label{subsec:cb_policy_eval}

Heterogeneity in treatment effects is a necessary condition for effective personalization, but the existence of HTEs does not necessarily imply that what is best for the individual student differs from what is best for the overall student population. Comparing panels (B) and (C) of Figure \ref{fig:conceptual} one can observe that a contextual policy that considers effect heterogeneity can only improve outcomes over a non-contextual (MAB) policy if the CATE function is sign-changing--i.e., depending on the student covariate the expected effect of the treatment can be negative and positive.

More formally, an assistance policy is a function $\pi: \mathbb{R}^d \rightarrow \{0, 1\}$ mapping student covariate vectors to treatment decisions. The value of a policy $\pi$ is defined as $v(\pi) \defeq \mathbb{E}[Y(\pi(x))]$. In the space of possible policy functions $\Pi$ the optimal policy is $\pi^* \defeq \arg \max_{\pi \in \Pi} v(\pi)$. Provided CATE function $\tau: \mathbb{R}^d \rightarrow \mathbb{R}$, we can characterize the optimal policy as
\begin{equation}
\pi^*(x) = \begin{cases} 
1 & \text{if } \tau(x) > 0 \\
0 & \text{otherwise}
\end{cases}
\end{equation}
which implies,
\begin{equation}
    \tau(x) > 0, \,\forall x \in \mathbb{R}^d \implies \pi^*(x) = 1, \forall x \in \mathbb{R}^d \implies v(\pi^*) = \mathbb{E}[Y(1)]
\end{equation}
and analogously
\begin{equation}
    \tau(x) < 0, \,\forall x \in \mathbb{R}^d \implies \pi^*(x) = 0, \forall x \in \mathbb{R}^d \implies v(\pi^*) = \mathbb{E}[Y(0)].
\end{equation}
Thus, non-contextual policies that optimize for ATE and make the same decision for all students achieve optimal outcomes whenever the CATE function is \textit{not} sign-changing. To assess whether assistance policies that consider student covariates can improve outcomes over non-contextual policies we implement the following procedure: First, for each question in our dataset we identify the assistance action that yields optimal outcomes for the overall student population. These actions resemble the MAB policies discussed in Section \ref{subsec:offline_evaluation}. Second, we construct action-specific datasets to study causal effects associated with the decision of providing that particular action (treatment) over the MAB action that is optimal for the overall student population (control). Third, for each dataset we train an estimator $\hat{\tau}$ predicting the CATE function using the CRF algorithm. Using estimator $\hat{\tau}$ we define an approximation of the optimal contextual policy as
\begin{equation}
\hat{\pi}(x) = \begin{cases} 
1 & \text{if } \hat{\tau}(x) > 0 \\
0 & \text{otherwise.}
\end{cases}
\end{equation}
In other words, the above approximation to the optimal contextual policy selects, for each value $x$ of student covariates, either the action recommended by the MAB policy (i.e., 0), or the alternative action (i.e., 1), depending on which is predicted by the CATE function to have the best learning outcome. Note that $\hat{\pi}$ will only differ from the MAB policy if  there is an heterogeneous effect (i.e., if  there exists at least one student covariate vector where the MAB is optimal and a second covariate vector where the alternative action is optimal). Fourth, we assess whether $\hat{\pi}$ improves learning outcomes over the MAB policy by estimating policy values and testing for significant differences via a one-sided $t$-test. By implementing this methodology for each individual action and outcome measure, we aim to evaluate the degree to which contextual assistance policies that account for individual student differences can yield significant improvements in outcomes over non-contextual MAB policies.

\section{Results}
\label{sec:results}

We present results in three parts: First, we estimate the effects of individual assistance actions on different measures of learning outcomes. Second, we train and evaluate multi-armed bandit policies that optimize instructional decision making \textit{on average} across the student population. Third, we evaluate the potential of contextual bandit policies that optimize instructional decision making for the \textit{individual} students.

\subsection{Evaluation of Assistance Actions}

We estimate the effects of \textit{individual} assistance actions on different measures of student learning outcomes by leveraging the log data collected by the randomized assistance policy (Table \ref{tab:offline_evaluation}). This allows us to quantify for each question how the different ways of supporting students with feedback after an incorrect response impact students' overall practice experience.


\begin{figure}[!htbp]
\centering
\begin{mdframed}[roundcorner=10, linecolor=myblue, leftmargin=26pt, userdefinedwidth=0.94\columnwidth]
\centering
(A) Practice Question
\vspace{1mm}
\begin{mdframed}[roundcorner=10, linecolor=black, leftmargin=0pt]
\centering
\includegraphics[width=0.8\linewidth]{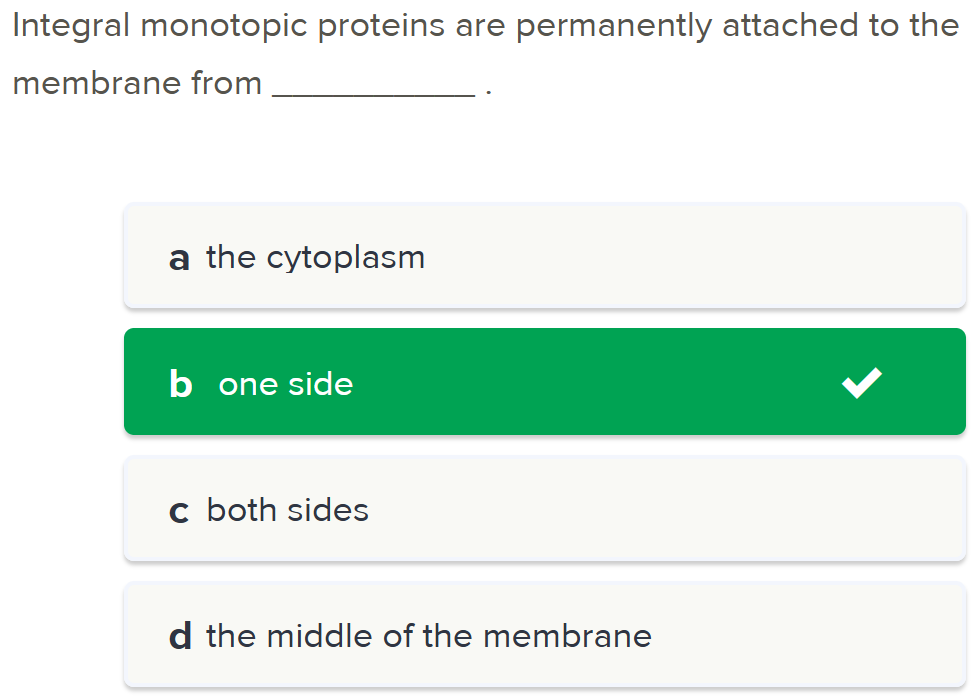}
\end{mdframed}
(B) Assistance Actions
\vspace{1mm}
\begin{mdframed}[roundcorner=10, linecolor=black, leftmargin=0pt]
\noindent $\bullet$ \textit{hint 1}: The prefix \textbf{mono}- means \textbf{single} or \textbf{alone}.
\vspace{1.5mm}

\noindent $\bullet$ \textit{hint 2}: The word \textbf{integral} means \textbf{essential} or \textbf{necessary to make whole}.
\vspace{1.5mm}

\noindent $\bullet$ \textit{paragraph}: Transmembrane proteins span the entire plasma membrane. Transmembrane proteins are found in all types of biological membranes. Integral monotopic proteins are permanently attached to the membrane from only one side.
\vspace{1.5mm}

\noindent $\bullet$ \textit{vocabulary}: \\
\small{
\indent - cytoplasm: Material inside the cell membrane, including the watery cytosol and other cell structures except the nucleus if one is present. \\
\indent - protein: A peptide that is greater than one hundred amino acids in length.}
\end{mdframed}
(C) Assistance Action Performance Estimates\\
\vspace{1mm}
\resizebox{0.95\columnwidth}{!}{
\begin{tabular}{lrrrrr} 
\hline
& n & Reward & Reatt. Cor. & Stud. Abil. & Sess. Succ. \\ 
\hline
no assist. & 872 &  0.025 ±.079 & 0.313 ±.031 & -0.167 ±.125 & 0.925 ±.017 \\
hint 1     & 986 &  \textbf{0.394} ±.077 & 0.693 ±.029 &  \textbf{0.195} ±.121 & 0.934 ±.015 \\
hint 2     & 940 & -0.029 ±.081 & 0.255 ±.028 & -0.219 ±.130 & 0.920 ±.017 \\
paragraph  & 940 &  0.363 ±.078 & \textbf{0.774} ±.027 &  0.088 ±.121 & \textbf{0.935} ±.016 \\
vocabulary & 936 & -0.033 ±.078 & 0.259 ±.028 & -0.228 ±.126 & 0.929 ±.016 \\
\hline
\end{tabular}}
\\
\phantom{phantom} \\ \vspace{-2.5mm}
\end{mdframed}
\caption{Example of assistance action evaluation for one individual question. By showing different assistance actions to different students we can quantify the effects of each assistance action on various measures of learning outcomes, as summarized in panel (C).}
\label{fig:example_question_evaluation}
\end{figure}

One example of the results of this evaluation process is illustrated by Figure \ref{fig:example_question_evaluation}. The figure shows question text, the set of available assistance actions as well as estimates of how each individual action affects different outcome measures. We observe that the \textit{paragraph} that provides detailed information leads to the highest reattempt correctness rate. In comparison, \textit{hint 1} leads to a lower reattempt correctness rate, but conveys insights that yield better overall session performance as captured by the student ability score. Further, we can identify assistance actions that are not helpful to students. For example, \textit{hint 2} and \textit{vocabulary} both provide information that is relevant for the question, but lead to worse outcome measures than showing \textit{no assistance}. Overall, these estimates are very compelling for the content creators, as they allow them to reflect on how the individual learning resources they designed affect the student learning experience. Using this data, the designers can identify questions where assistance has large effects on learning outcomes, and they can identify cases where assistance content should be revised. 

\begin{figure*}[t]
    \centering
    \includegraphics[width=0.495\textwidth]{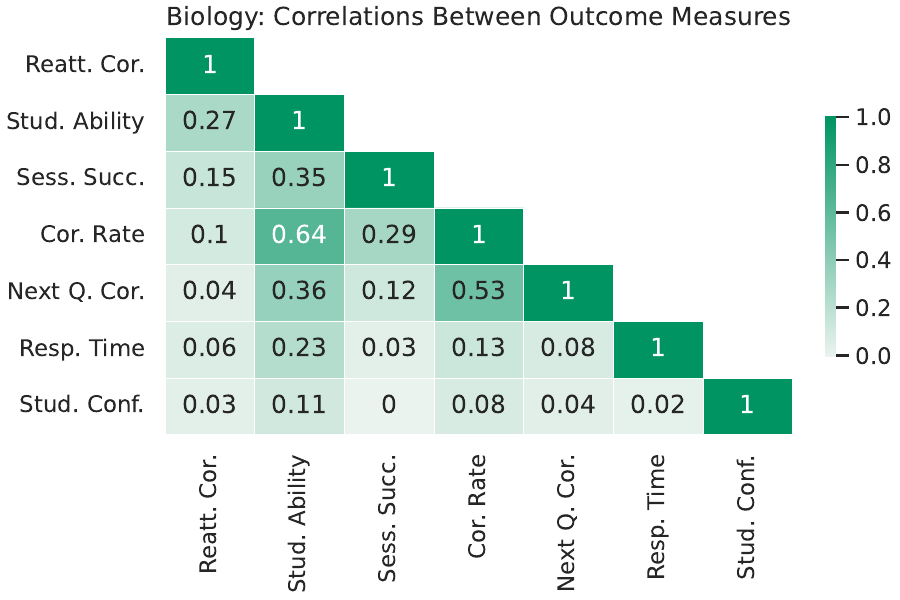}
    \hfill
    \includegraphics[width=0.495\textwidth]{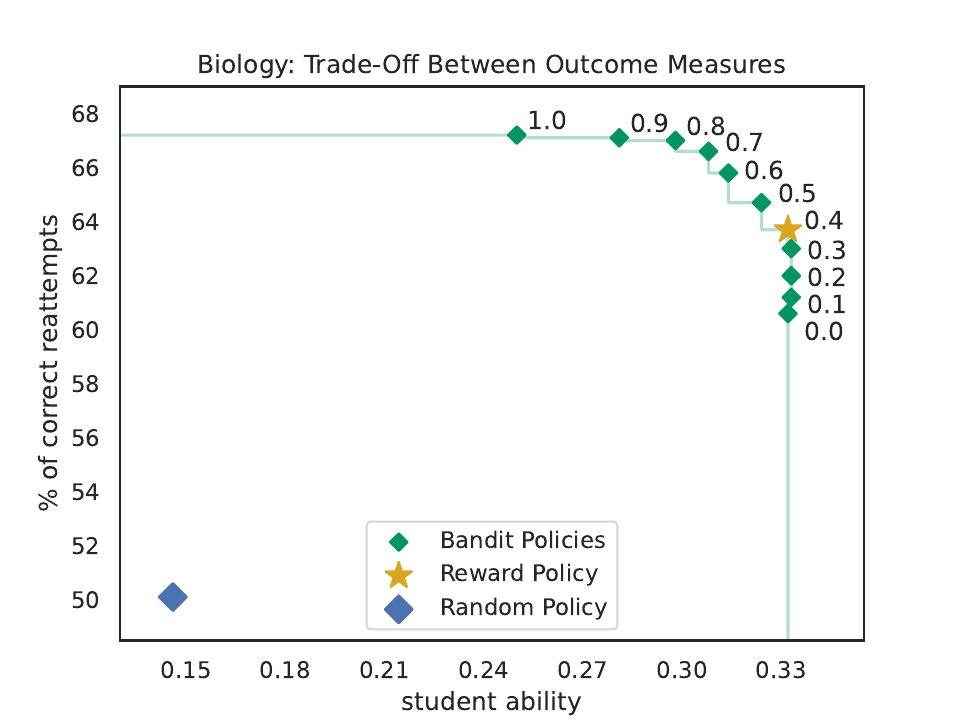}
    \caption{[Left] Average within question correlations between different measures of learning outcomes across $1,336$ questions. From top to bottom, the considered measures are reattempt correctness, student ability, session success, future correct rate, next question correct, future response time, and student confidence. Recall that "student ability" refers to the ability estimate at the end of the session, including over questions asked after the teaching action was taken. [Right] Pareto front visualizing the estimated average performance of multi-armed bandit policies optimized to increase the final student ability estimates ($x$-axis) and reattempt correctness rates ($y$-axis) across $178$ questions. Each bandit policy is marked with a number that indicates how it weights the two objectives (see Equation~\ref{eq:reward}).
    }
    \label{fig:signal_correlations}
\end{figure*}

To gain insight into the relationships between the different learning outcome measures we analyse average within question correlations across the $1,336$ biology questions (Figure \ref{fig:signal_correlations}). The focus on within question correlation makes our analyses robust towards effects caused by systematic differences between individual questions (e.g., due to varying difficulty levels). We find reattempt correctness rate to be most correlated with the IRT-based student ability estimates ($r = 0.27$) and mostly uncorrelated with next question correctness ($r = 0.04$). This highlights that while assistance actions can improve students' overall session performance, due to differences between individual questions, one needs to consider more than just the next question. Matching our intuition, we observe that student ability estimates are correlated with other performance indicators including session success ($r = 0.35$), future correctness ($r=0.64$) and next question correctness rates ($r = 0.36$) which all consider first attempt response correctness. Student response time exhibits a low positive correlation to student ability ($r = 0.23$) which might be due to different problem solving strategies (e.g., some students rely more on assistance actions). We find that the self-reported student confidence measure shows very low correlations with other measures. This might be due to the system's IRT-based question sequencing strategy that assigns more difficult questions to students exhibiting higher performance levels.

\begin{figure*}[t]
    \includegraphics[width=1.0\textwidth]{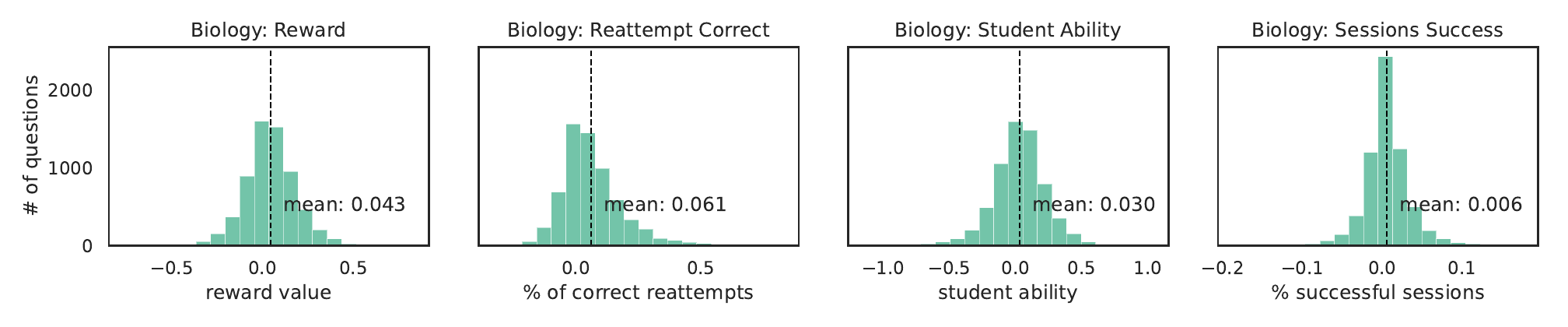}
    \caption{Histogram visualizing effect differences between individual assistance actions and \textit{no assistance} baselines for different measures of learning outcomes for each of the $1,336$ biology questions. On average each question is associated with $5.8$ different actions.}
    \label{fig:hist_performance_gap}
\end{figure*}

To gauge the degree to which assistance policies can impact the student learning experience we estimate the treatment effects of \textit{individual} assistance actions in the biology course by comparing the outcomes of students that received a particular assistance action to students that received \textit{no assistance} (Figure \ref{fig:hist_performance_gap}). Across the $1,336$ biology questions we observe an average treatment effect of $6.1\%$ on reattempt correctness rates, of $0.030$ on student ability estimates and of $0.6\%$ on practice session success rates. Even though the assistance content was curated by human domain experts, we find that a substantial number of actions do not perform better than the \textit{no assistance} baseline. This underscores the need for data-driven evaluations and shows that learned assistance policies have the potential to improve learning outcomes significantly.

Lastly, before optimizing assistance policies we study the degree to which the available student log data (Table \ref{tab:data_overview}) allows us to differentiate between the effects of assistance actions for individual questions via analysis of variance (ANOVA). Compared to the multi-armed bandit problem which focuses on identifying the single most effective action, ANOVA asks the simpler question of whether there are statistically significant differences in mean effects between the available assistance actions--a precondition for any successful intervention. On average across the six science courses, ANOVA rejects the null hypothesis ($p < 0.05$) for \textit{reattempt correctness} for $70.5\%$, for \textit{student ability} for $11.0\%$ and for \textit{session success} for $10.9\%$ of practice questions. The exact rejection rates for the individual courses are provided in Table \ref{tab:anova}. First, we observe that courses with more available assistance content (e.g., biology and earth science) have higher rejection rates compared to courses with less content (e.g., chemistry and physics) highlighting that a rich base of assistance content is a precondition for learning effective assistance policies (compare Table \ref{tab:data_overview}). Second, we observe that ANOVA rejects the null hypothesis significantly more often for \textit{reattempt correctness} compared to \textit{student ability} and \textit{session success} measures. This can be explained by studying the relationship between sample variance and effect size gaps between the most and least effective assistance action for the different measures. Formally, for each question $q$ we define the effect size gap as $\delta_q \defeq \max_{a_q \in A_q} \mathbb{E}[r_{a_q}] - \min_{a_q \in A_q} \mathbb{E}[r_{a_q}]$ where $r_{a_q}$ is a reward measure of interest (e.g., reattempt correctness). On average across the $1,336$ biology questions, \textit{reattempt correctness} has better ratio between action effect gaps and sample variance ($\delta = 0.230$, $\sigma^2 = 0.221$) compared to \textit{student ability} ($\delta = 0.302$, $\sigma^2 = 3.621$) and \textit{session completion} ($\delta = 0.042$, $\sigma^2 = 0.084$). This is because \textit{reattempt correctness} focuses on students' immediate interactions with the current question, while the two latter measures gauge overall session performance over a longer time horizon. For our goal of learning effective assistance policies, this implies that one requires more data to learn policies that optimize students' long-term performance compared to  short-term outcomes.


\begin{table}[t]
\centering
\caption{Proportion of questions for which ANOVA detects significant differences ($p < 0.05$) between the effects of assistance actions in our content pool for different measures of learning outcomes.}
\begin{tabular}{lrrrrrr}
\hline
 measure/subject & Biology & Chemistry & Physics & Life Sci. & Earth Sci. & Phys. Sci.\\ 
\hline
Reatt. Cor. & 83.2\% & 61.8\% & 50.5\% & 75.9\% & 74.2\% & 77.5\% \\   
Stud. Abil. & 13.3\% &  8.5\% &  8.2\% & 11.8\% & 11.6\% & 12.9\% \\
Sess. Succ. &  9.6\% &  7.9\% & 10.5\% & 11.2\% & 13.3\% & 13.2\% \\
\hline
\end{tabular}
\label{tab:anova}
\end{table}



\subsection{Evaluation of Multi-Armed Bandit Policies}
\label{subsec:bandit_results}

\subsubsection{Offline Policy Evaluation}


\begin{table}[t]
\centering
\caption{Offline evaluation of different multi-armed bandit policies across the $178$ biology questions for which ANOVA indicated significant differences ($p < 0.05$) in mean action effects on student ability. The first two rows are two baselines policies, one which does not show assistance at all and one which selects assistance actions for each question at random. The following three rows are multi-armed bandit policies trained to optimize different measures of learning outcomes. The last row is the policy learned using our combined reward function which considers reattempt correctness and student ability. We report mean values and $95\%$ confidence intervals.}
\begin{tabular}{lrrrr}
\hline
policy/measure& Reward & Reatt. Cor. & Stud. Abil. & Sess. Succ. \\ 
\hline
random & 0.288 ±.067 & 0.501 ±.021 & 0.146 ±.109 & 0.806 ±.039 \\ 
\hline
reattempt correct & 0.419 ±.071 & \textbf{0.672} ±.020 & 0.250 ±.115 & 0.816 ±.038 \\ 
student ability & 0.442 ±.068 & 0.606 ±.025 & \textbf{0.332} ±.109 & 0.819 ±.038 \\ 
session success. & 0.371 ±.068 & 0.573 ±.025 & 0.237 ±.109 & 0.817 ±.037 \\
reward & \textbf{0.454} ±.068 & 0.637 ±.023 & \textbf{0.332} ±.108 & \textbf{0.820} ±.038 \\
\hline
\end{tabular}
\label{tab:offline_evaluation_anova}
\end{table}


\begin{table}[t]
\centering
\caption{Offline evaluation of different multi-armed bandit policies across $1,336$ biology questions using data from $3,266,171$ instances where assistance was provided. The first two rows are two baselines policies, one which does not show assistance at all and one which selects assistance actions for each question at random. The following three rows are multi-armed bandit policies trained to optimize different measures of learning outcomes. The last row is the policy learned using our combined reward function which considers reattempt correctness and student ability. We report mean values and $95\%$ confidence intervals.}
\begin{tabular}{lrrrr}
\hline
policy/measure & Reward & Reatt. Cor. & Stud. Abil. & Sess. Succ. \\ 
\hline
random & 0.255 ±.026 & 0.551 ±.007 & 0.058 ±.042 & 0.820 ±.013 \\ 
\hline
reattempt correct & 0.327 ±.026 & \textbf{0.666} ±.007 & 0.101 ±.043 & \textbf{0.827} ±.013 \\
student ability & 0.327 ±.026 & 0.660 ±.007 & \textbf{0.105} ±.043 & \textbf{0.827} ±.013 \\ 
session success & 0.326 ±.026 & 0.663 ±.007 & 0.101 ±.043 & \textbf{0.827} ±.013 \\
reward & \textbf{0.328} ±.026 & 0.664 ±.007 & 0.104 ±.043 & \textbf{0.827} ±.013 \\ 
\hline
\end{tabular}
\label{tab:offline_evaluation}
\end{table}

While ANOVA finds significant differences in mean action effects on \textit{reattempt correctness} for most questions, it only detects differences in \textit{student ability} and \textit{session completion} for a smaller subset of questions (Table \ref{tab:anova}). For our offline policy evaluation process this suggests that it is difficult to reliably identify the optimal assistance actions for the latter two measures even when having access to hundreds of samples per action. Indeed, in preliminary experiments we found that action effect rankings based on training data often deviate from rankings based on separate test data. For the average  question we found training assistance policies based on \textit{reattempt correctness} estimates to be the most effective way to boost all three outcome measures. This is due to its lower variance and the fact that improvements in \textit{reattempt correctness} are positively correlated with improvements in \textit{student ability} and \textit{session completion rates} (Figure \ref{fig:signal_correlations} left).

Still, for $11\%$ of questions ANOVA detected significant differences in action effects on \textit{student ability}, which is a central measure of interest--it captures students' overall session performance, including responses to questions \textit{after} the provided assistance. For example, in the biology course we detected significant differences in student ability based on which assistance action they received, for $178$ ($13.3\%$) of the 1,366 questions. To study the relationship between \textit{reattempt correctness} and \textit{student ability}, we train bandit policies for different objectives for these $178$ questions. In analogy to the reward function (Equation \ref{eq:reward}), we assign each policy a weight $w_1 \in \{0, 0.1, \dots, 1.0\}$ and compute its reward values by linearly weighting \textit{reattempt correctness} with $w_1$ and \textit{student ability} with $1 - w_1$. We visualize the Pareto front defined by the resulting policies (Figure \ref{fig:signal_correlations} right) and observe performance estimates that range in \textit{reattempt correctness} rates from $60.6\%$ to $67.2\%$ and in \textit{student ability} from $0.250$ to $0.332$. All learned policies outperform the \textit{random} policy significantly. In collaboration with domain experts we select $w_1 = 0.4$ as reward function to train the assistance policy for live evaluation as it improves both measures substantially. Table \ref{tab:offline_evaluation_anova} provides detailed performance statistics for policies trained to optimize different outcome measures.

To train an assistance policy for all questions we designed an algorithm that for each question decides whether we have sufficient data to optimize the measure of interest (i.e., \textit{reward}) directly or whether we should use the lower variance \textit{reattempt correctness} measure (details in Section \ref{sec:methodology}). Relatedly, Table \ref{tab:offline_evaluation} shows average performance metrics across the $1,336$ biology questions for a policy that always selects the \textit{no assistance} action, the \textit{random} policy, and four policies trained using our algorithm to optimize \textit{reattempt correctness} rates, \textit{student ability}, \textit{successful session completion} rates, and \textit{reward} function respectively. The algorithm resolves the variance issue and the trained policies enhance the student experience in different ways.

Lastly, we study for which types of questions the final policy selects which types of assistance actions to maximize the \textit{combined reward} objective. Table \ref{tab:biology_action_dist} shows for each question type for what proportion of questions the policy finds a certain assistance type to be most effective. We find that the policy utilizes a diverse blend of assistance types for each type of question and that paragraph actions are selected most frequently overall. Because of this, we compared the effects of a policy that always selects \textit{paragraph} actions to the trained \textit{reward} policy in an additional experiment. Across the $1,175$ biology questions with paragraphs, we find that the \textit{reward policy} outperforms the \textit{paragraph} policy in all measures (\textit{reward:} $0.336$/$0.299$, \textit{reattempt correctness:} $67.3\%$/$61.5\%$, \textit{student ability:} 0.112/0.089, \textit{session success:} $84.0\%$/$83.7\%$). The data-driven approach thus benefits the tutoring system by selecting effective teaching actions on a question-by-question basis.


\begin{table}[t]
\centering
\caption{Types of assistance actions selected by the multi-armed bandit policy learned using our reward function. The individual columns show how the policy focuses on different types of assistance actions for different types of questions in the biology course.}
\begin{tabular}{lrrrr}
\hline
action/question\,\, & Mult.-Choice & All-That-Apply  & Fill-Blank & Short-Answ. \\ 
\hline
no assistance   &  5.0\% &  7.3\% &  1.8\% &  3.4\% \\
hint         & 11.4\% & 15.6\% &  6.6\% &  5.6\% \\
paragraph    & 51.2\% & 43.1\% & 57.9\% & 55.1\% \\
vocabulary   &  5.7\% & 16.5\% &  1.8\% &  3.4\% \\
hide distractor  & 26.8\% & 17.4\% &      - &      - \\
first letter &      - &      - & 31.7\% & 32.6\% \\
\hline
\end{tabular}
\label{tab:biology_action_dist}
\end{table}


\begin{table}[t]
\centering
\setlength{\tabcolsep}{4pt}
\caption{Live policy evaluation. We randomly assign student practice sessions to the randomized policy and the learned multi-armed bandit policy condition, track various outcome measures and report mean values and $95\%$ confidence intervals. Overall, $34,279$ students were assigned to the randomized policy and $33,957$ students were assigned to the multi-armed bandit policy.}
\begin{tabular}{lcccccc}
\hline
Subject & Policy & Num. Sessions & Reward & Reatt. Cor. & Stud. Abil. & Sess. Succ. \\ 
\hline
\multirow{2}{*}{Life Sci.} & random & 6,655 & 0.783 ±.037 & 0.559 ±.008 & 0.933 ±.059 & 0.608 ±.012 \\
& learned & 6,831 & \textbf{0.933} ±.035 & \textbf{0.642} ±.008 & \textbf{1.126} ±.056 & \textbf{0.651} ±.011 \\
\hline
\multirow{2}{*}{Earth Sci.} & random & 9,559 & 0.576 ±.030 & 0.593 ±.007 & 0.566 ±.048 & 0.441 ±.010 \\
& learned & 9,557 & \textbf{0.694} ±.030 & \textbf{0.658} ±.007 & \textbf{0.718} ±.047 & \textbf{0.476} ±.010 \\
\hline
\multirow{2}{*}{Phys. Sci.} & random & 10,035 & 0.645 ±.028 & 0.552 ±.007 & 0.707 ±.045 & 0.509 ±.010 \\
& learned & 10,208 & \textbf{0.706} ±.027 & \textbf{0.626} ±.007 & \textbf{0.759} ±.043 & \textbf{0.535} ±.010 \\
\hline
\multirow{2}{*}{Biology} & random & 31,527 & 0.753 ±.016 & 0.585 ±.004 & 0.866 ±.026 & 0.676 ±.005 \\
& learned & 30,937 & \textbf{0.881} ±.015 & \textbf{0.683} ±.004 & \textbf{1.013} ±.024 & \textbf{0.712} ±.005 \\
\hline
\multirow{2}{*}{Chemistry} & random & 21,689 & 0.813 ±.020 & 0.547 ±.005 & 0.991 ±.032 & 0.581 ±.007 \\
& learned & 21,675 & \textbf{0.882} ±.020 & \textbf{0.592} ±.005 & \textbf{1.075} ±.031 & \textbf{0.604} ±.007 \\
\hline
\multirow{2}{*}{Physics} & random & 3,871 & 0.894 ±.048 & 0.528 ±.011 & 1.139 ±.077 & 0.475 ±.016 \\
& learned & 3,816 & \textbf{0.946} ±.049 & \textbf{0.568} ±.012 & \textbf{1.197} ±.078 & \textbf{0.486} ±.016 \\
\hline
\end{tabular}
\label{tab:online_policy_evaluation}
\end{table}

\subsubsection{Real-World Policy Evaluation}
To evaluate the policies optimized using our training algorithm and combined reward function we compare their ability to provide students with effective assistance to the randomized assistance policies. The initial A/B evaluation centered on the high school biology course (April 5th to April 13th, 2023) featuring a large diversity of assistance actions. In a second step we expanded the evaluation to the other five science courses (June 7th to July 10th, 2023). In both evaluations practice sessions were randomly assigned either to the bandit policy or to the randomized policy condition. Overall, we collected log data from over 166,000 student practice sessions. 

Table \ref{tab:online_policy_evaluation} reports average outcome measures achieved by the different assistance policies for each of the six courses. The trained assistance policies outperform the randomized policies in all subjects in all outcome measures, achieving for individual courses average improvements in reattempt correctness rates between $5.2\%$ and $15.0\%$, in student ability estimates between $0.052$ and $0.193$, and in session success rates between $1.1\%$ and $4.3\%$. In biology, the session success rate improvement from $67.6\%$ to $71.2\%$ corresponds to a $11.1\%$ reduction in sessions in which students did not achieve the practice target. We observe that the optimized policies are particularly effective in subjects featuring a larger variety of assistance content (Table \ref{tab:data_overview}). We note that in contrast to the prior offline evaluation experiments where we estimate effects based on individual assistance queries, here we compute metrics based on the session level.

\subsection{Evaluation of Contextual Bandit Policies}

\subsubsection{Offline Heterogeneous Treatment Effect Evaluation}

\begin{figure}
\centering
%
\begin{minipage}{0.65\textwidth}
\vfill
\textbf{Question}: The exoskeleton of a grasshopper is made up of what? a) glycogen; b) \underline{chitin}; c) cellulose; d) starch\\

\vspace{-2mm}
\noindent \textbf{Assistance}:
\textit{cellulose}: complex carbohydrate that is a polymer of glucose and that makes up the cell wall of plants; \textit{chitin}: tough carbohydrate that makes up the cell walls of fungi and the exoskeletons of insects and other arthropods; \textit{exoskeleton}: a hard covering that supports and protects an animal's body; \textit{starch}: large, complex carbohydrate found in foods such as grains and vegetables that the body uses for energy.
\end{minipage}\hfill
\begin{minipage}{0.35\textwidth}
\centering
\hfill \includegraphics[width=0.90\linewidth]{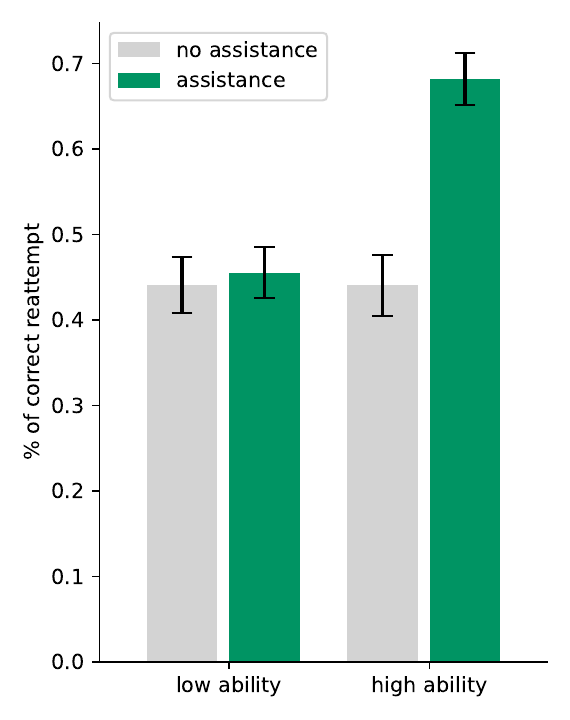}
\end{minipage}
 
\noindent \begin{minipage}{0.65\textwidth}
\vfill
\noindent \textbf{Question}: What is a wave in which particles of matter vibrate parallel to the direction the wave travels called? \underline{longitudinal wave} (short-answer)\\

\vspace{-2mm}
\noindent \textbf{Assistance}: In an earthquake, what type of wave is a P wave?
\end{minipage}\hfill
\begin{minipage}{0.35\textwidth}
\centering
\hfill \includegraphics[width=0.90\linewidth]{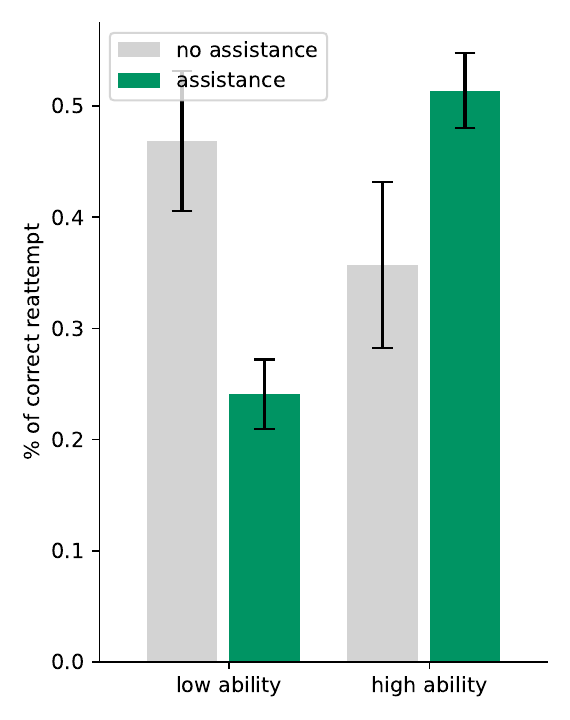}
\end{minipage}

\noindent \begin{minipage}{0.65\textwidth}
\vfill
\noindent \textbf{Question}: Santa Cruz Island off the California coastline includes steep cliffs, coves, gigantic caves, and sandy beaches. What are these physical environments where organisms live known as? a) niche; b) \underline{habitat}; c) ecology; d) biotic factors\\

\vspace{-2mm}
\noindent \textbf{Assistance}: \textit{ecology}: branch of biology that studies how living things interact with each other and with their environment; \textit{niche}: an organism’s “job” within its community; \textit{organism}: an individual living thing; \textit{habitat}: physical environment in which a species lives and to which it has become adapted.
\end{minipage}\hfill
\begin{minipage}{0.35\textwidth}
\centering
\hfill \includegraphics[width=0.90\linewidth]{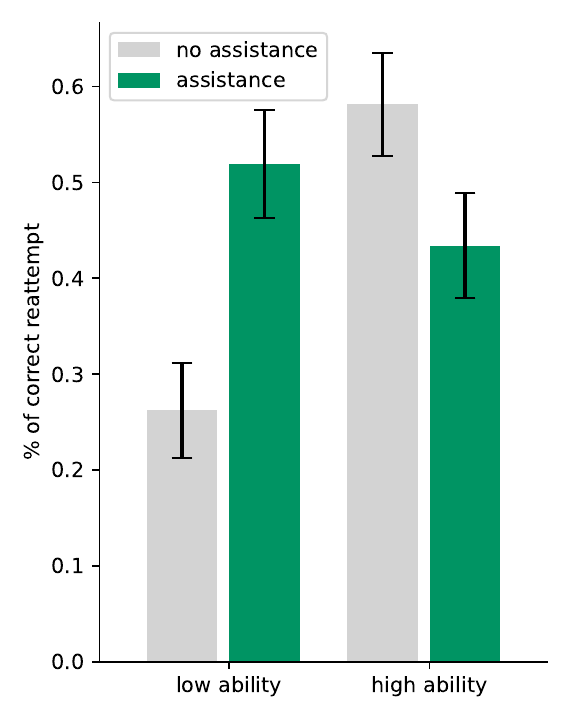}
\end{minipage}
\caption{Examples of heterogeneous treatment effects (HTEs) of assistance actions on reattempt correctness outcomes. Each plot compares learning outcomes of students receiving a particular assistance action,  versus students receiving the ``no assistance'' baseline. We assign students into ``high'' and ``low'' ability groups based on whether their current IRT ability estimate places them above or below the median value, then estimate reattempt correctness rates for each group. Note in each case, the assistance action produces significantly different learning outcomes across the two groups. }
\label{fig:hte_examples}
\end{figure}

We begin by presenting findings from hypothesis tests detecting heterogeneous treatment effects (HTEs) in binary decisions about providing students with a particular assistance action (treatment) or ``no assistance'' (control). HTEs are a precondition for personalization to be effective (e.g., contextual bandits), because without HTEs what is best for the overall student population is also best for the individual student. Following the methodology described in Section \ref{subsec:contextual_methodology} we test for linear and non-linear HTEs. For illustrative purposes, we provide some examples of assistance actions for which effects on reattempt correctness outcomes vary across student groups whose IRT ability estimates are above/below the median (Figure \ref{fig:hte_examples}). 

\textbf{Linear HTE} For each assistance action, learning outcome measure and student covariate (details Table \ref{tab:feature_overview}) we perform regression analysis to assess linear interactions between treatment effect and student covariate. Table \ref{tab:linear_heterogeneity} reports average detection rates across $7,548$ individual assistance actions for each outcome-covariate combination. The position of the current question in the learning process (quest. num.) is the covariate associated with the highest HTE detection rates for reward ($9.8\%$), student ability ($10.3\%$) and session success ($0.4\%$) outcome measures. The covariate with second highest HTE detection rates is the student's IRT ability estimate at time of first attempt ($7.3\%$ for reward and $8.0\%$ for student ability outcomes). Student ability outcomes are an aggregated measure of student performance. Time of assistance provision as well as current student ability likely modulate the degree to which future responses can impact the final IRT ability outcome. We detect little HTE for reattempt correctness and session success outcome measures where the covariates with highest detection rates are second attempt correctness rate ($1.1\%$) and question number ($0.4\%$).


\begin{table}[t]
\centering
\caption{Study of Linear Heterogeneous Treatment Effects (HTEs). For each of $7,458$ assistance actions, we fit regression models predicting outcomes of students receiving assistance (treatment) and students receiving ``no assistance'' (control). For each covariate/outcome combination, the regression model fits three parameters capturing average treatment effect (ATE), student covariate effect and heterogeneous treatment effect (HTE). We report the proportion of actions for which significant HTE was detected ($p < 0.05$, FDR controlled at $0.2$).}
\begin{tabular}{lrrrr}
\hline
covariate/measure & Reward & Reatt. Cor. & Stud. Abil. & Sess. Succ. \\
\hline
stud. ability & 7.3\% & 0.9\% & 8.0\% & 0.0\% \\
resp. time & 0.3\% & 0.1\% & 0.3\% & 0.0\% \\
prev. resp. cor. & 0.8\% & 0.0\% & 0.6\% & 0.2\% \\
quest. num & \textbf{9.8}\% & 0.3\% & \textbf{10.3}\% & \textbf{0.4}\% \\
cor. rate  & 1.8\% & 0.1\% & 1.8\% & 0.0\% \\
\hline
assigned   & 0.0\% & 0.0\% & 0.0\% & 0.0\% \\
confidence & 0.0\% & 0.0\% & 0.0\% & 0.0\% \\
weekend    & 0.0\% & 0.0\% & 0.0\% & 0.0\% \\
\hline
num sess. total   & 0.2\% & 0.0\% & 0.2\% & 0.0\% \\
num quest. total  & 0.4\% & 0.0\% & 0.4\% & 0.0\% \\
num assist. total & 1.4\% & 0.0\% & 1.6\% & 0.0\% \\
\hline
avg quest. num  & 5.6\% & 0.0\% & 6.2\% & 0.0\% \\
avg sess. succ. & 0.1\% & 0.1\% & 0.2\% & 0.0\% \\
\hline
avg 1st cor. & 2.5\% & 0.7\% & 2.1\% & 0.0\% \\
avg 2nd cor. & 0.5\% & \textbf{1.1}\% & 0.4\% & 0.0\% \\
\hline
avg 1st assists. & 0.0\% & 0.0\% & 0.0\% & 0.0\% \\
avg 2nd assists. & 0.2\% & 0.0\% & 0.3\% & 0.0\% \\
\hline
med 1st resp. time & 0.6\% & 0.0\% & 0.4\% & 0.1\% \\
med 2nd resp. time & 2.5\% & 0.0\% & 2.9\% & 0.1\% \\
med assist. time   & 0.7\% & 0.0\% & 0.4\% & 0.0\% \\
\hline
\end{tabular}
\label{tab:linear_heterogeneity}
\end{table}

\textbf{Non-Linear HTE} We test for non-linear HTE via CRF-based residual and RATE analysis. Unlike the regression approach which tests for HTE considering one student covariate at a time, CRF estimates the CATE function using complete covariate vectors as input. Table \ref{tab:nonlinear_heterogeneity} reports average HTE detection rates for CRF-based residual and RATE analysis across $7,458$ assistance actions. The residuals analysis rejects the null hypothesis for $0.2\%$ of actions for reward and student ability measures and for $2.1\%$ of actions for reattempt correctness. RATE analysis yields rejection rates of $0.2\%$, $0.1\%$ and $0.8\%$ for reward, student ability and reattempt correctness outcomes respectively. For the session success outcome measure, residual analysis detected HTE for one action and RATE analysis detected HTE for three actions.


\begin{table}[t]
\centering
\caption{Study of Non-Linear Heterogeneous Treatment Effects (HTEs). For each of $7,458$ assistance actions, we fit causal random forest (CRF) models predicting outcomes of students receiving assistance (treatment) and students receiving ``no assistance'' (control). We report the proportion of actions for which significant HTE was detected ($p < 0.05$, FDR   controlled at $0.2$) based on residual and rank-weighted average treatment effect (RATE) analysis.}
\begin{tabular}{lrrrr}
\hline
test/measure & Reward & Reatt. Cor. & Stud. Abil. & Sess. Succ. \\
\hline
Residual Analysis & 0.2\% & 2.1\% & 0.2\% & 0.0\% \\
RATE & 0.2\% & 0.8\% & 0.1\% & 0.0\% \\
\hline
\end{tabular}
\label{tab:nonlinear_heterogeneity}
\end{table}

\subsubsection{Offline Policy Evaluation}


\begin{table}[t]
\centering
\caption{Number of assistance actions for which contextual bandit policies leveraging HTEs showed significant improvements in learning outcomes over multi-armed bandit policies.}
\begin{tabular}{cccc}
\hline
Reward & Reatt. Cor. & Stud. Abil. & Sess. Succ. \\
\hline
0 & 3 & 1 & 0 \\
\hline
\end{tabular}
\label{tab:policy_comparison}
\end{table}

To assess potential benefits of personalized assistance selection we study effects of assigning individual students assistance actions (treatment) that deviate from what is best for the overall student population (control). In particular, we test whether contextual policies defined via CRF-based CATE function estimates achieve significant improvements in learning outcomes over non-contextual policies. In reinforcement learning terminology, this experiment corresponds to a comparison of contextual bandit and multi-armed bandit policies in a decision-problem with two actions.

Table \ref{tab:policy_comparison} reports the result of this experiment. Among all $7,548$ individual assistance actions, we detected significant benefits of personalization on reattempt correctness for three actions and on student ability for one action. The methodology did not reveal any actions for which personalization led to significant benefits for reward and session success outcome measures.

\section{Discussion}
\label{sec:discussion}

This paper presented a case study of data-driven design that optimized teaching policies for deciding which assistance action (e.g., one of multiple hints) to provide as feedback to students after they answer practice questions incorrectly. We analyzed a dataset of over $23.8$ million logs from one million students who interacted with $43,000$ randomly assigned assistance actions designed by human domain experts--the largest evaluation of feedback content to date. Using this dataset, we employed offline reinforcement learning algorithms to train multi-armed bandit (MAB) policies that optimize assistance actions for the \textit{overall student population} and contextual bandit (CB) policies that personalize feedback based on \textit{individual} student attributes. A live evaluation with over $166,000$ practice sessions demonstrated significant improvements in student outcomes compared to the data collection policy, confirming the system's ability to enhance its teaching using interaction log data. The following discusses novel insights, connections to prior research, study limitations, and future directions.

\textbf{Insights from Multi-Armed Bandit Policies}
We evaluated the effects of individual assistance actions on various measures of student learning outcomes, such as reattempt correctness and session success, and explored the relationships among these measures. By analyzing the actions selected by our optimized policies (Figure~\ref{tab:biology_action_dist}) we observe that there is no single assistance action format (e.g., hint, keyword definition) that is universally best for any type of question (e.g., multiple-choice, fill-in-the-blank). Preparing effective instruction is challenging, even for human domain experts~\citep{Nathan2001:Expert}, and our findings highlight the importance of data-driven algorithms that can identify effective teaching actions on a case-by-case basis.

Reinforcement learning enables a data-driven design process in which ITS designers specify a reward function to optimize teaching policies that promote desired learning outcomes. Our analysis identified a trade-off between assistance policies optimized for long-term outcomes (e.g., student ability) and short-term outcomes (e.g., reattempt correctness) (Figure~\ref{fig:signal_correlations} Right). To address this challenge, we developed an algorithm that determines the most suitable policy training objective for each question, enhancing students' success on their second attempt at the current question, as well as their overall performance during the practice session. Interestingly, the optimal policy blends more informative assistance actions (e.g., paragraphs) with less informative ones (e.g., hints) and, for some questions, opts to provide no additional help at all. This finding highlights a trade-off between providing and withholding information during the learning process, corroborating a phenomenon described as the \textit{assistance dilemma} in prior research~\citep{Koedinger2007:Exploring}.

\textbf{Insights from Contextual Bandit Policies} 
We investigated whether CB policies that personalize assistance by considering student features, such as current ability level and average response time, can enhance learning outcomes compared to the MAB policies currently deployed. Using statistical and causal machine learning, we developed a framework to (i) assess how assistance effects vary across individual students and (ii) evaluate whether CB policies leveraging these heterogeneous treatment effects (HTEs) can outperform MAB policies significantly. Specifically, we analyzed data from $420,000$ students collected in randomized experiments, and evaluated $7,548$ assistance actions considering various outcome measures (e.g., reattempt correctness, session success). While we identified significant variations in treatment effects for some actions (e.g., between low- and high-ability students), findings suggest that the overall magnitude of these effects were often insufficient for CB policies to significantly improve learning outcomes over non-contextual (MAB) approaches.

From a machine learning perspective, the implications of these findings within the confines of the studied tutoring system are that assistance policies optimized for the overall student population may, in many cases, also be optimal for individual students (Figure~\ref{fig:conceptual}). This suggests that MAB policies may achieve outcomes similar to those of CB policies, which make personalized instructional decisions. Compared to CB algorithms, MAB algorithms can be more data-efficient, particularly when context and outcome distributions are uncorrelated~\citep{Lattimore2020:Bandit}. Within the studied system, MAB algorithms may arrive at effective instructional policies faster than CB algorithms. Reflecting on our analysis, we identified various questions for which none of the available actions yielded significant outcome improvements over the ``no assistance'' baseline (Table~\ref{tab:anova}). Combined, these findings suggest that, rather than focusing on learning more complex policies, future iterations of the tutoring system are likely to benefit from a focus on feedback content refinements.

From a learning science perspective, the present work motivates future explorations targeted towards understanding how individual student characteristics affect sensitivity to different instructional conditions and interventions~\citep{Koedinger2013:Instructional}. Features of student behavior such as help-seeking~\citep{Aleven2003:Help} and gaming behavior~\citep{Baker2004:Off} are known to be predictive of learning outcomes, but how these features should inform instructional policies is still a topic of ongoing research. For example,~\citet{Nie2023:Understanding} employed offline evaluation to compare the effects of instructional policies that respond to users' questions by choosing among four support strategies (e.g., hinting and providing encouragement). Using data from $270$ participants, Nie et al. estimated the effects of a reinforcement learning policy making decisions based on student features (e.g., pre-test scores and math-anxiety levels) and a non-contextual policy to be similar. The large-scale dataset we collected will serve as foundation for future research exploring what are features of effective feedback elements and what are interactions between these features, individual student attributes and student learning outcomes~\citep{Aleven2016:Instruction, Bernacki2021:Systematic}.

\subsection{Limitations and Future Work}

The studied practice questions originated from six online science courses for middle and high school students. Despite this diversity, most questions within these courses can be classified as shallow forms of knowledge assessment focused on recall. Effect heterogeneity might be more prevalent in domains featuring deep understanding questions~\citep{Vanlehn2007:Tutorial}. Furthermore, despite the large scale of this study, we only considered a limited number of student covariates. Future work might consider additional factors, such as students' self-regulation strategies~\citep{Aleven2010:Automated}, mind-wandering~\citep{Mills2015:Mind}, and gaming behavior~\citep{Baker2004:Off}. While larger sample sizes can increase HTE detection rates, they are often unobtainable in most practical settings. In future work, we will explore whether all students in our dataset interacted with the provided assistance in a meaningful way~\citep{Aleven2003:Help}. If a student does not process the provided feedback on a cognitive level, their behavior may not be indicative of content quality, thus adding noise to our evaluations. Another direction is to study effect heterogeneity on an aggregated level, moving the focus from individual assistance actions to subsets of actions exhibiting similar characteristics--for example, by comparing hints to keyword definitions or comparing paragraphs with and without visual illustrations.

Our methodology optimized question-specific bandit policies in a two-stage process.  First, student log data was collected for each question using an evaluation policy that selected assistance actions uniformly at random. Second, after collecting a sufficient amount of data, offline policy evaluation was employed to train a policy that optimized our reward function. This process allowed us to evaluate the effects of different assistance policies prior to live deployments. In future work, we will implement online bandit algorithms~\citep{Lattimore2020:Bandit}, iteratively refining assistance policies by adaptively sampling available actions based on evolving effect estimates during live deployment. Adaptive sampling is of particular interest to us because the pool of questions and assistance actions within the tutoring system is continuously evolving. In this effort, we will need to consider that online bandit algorithms can exhibit higher false discovery rates compared to conventional randomized experiments~\citep{Rafferty2019:Statistical}. Another interesting direction for future research is whether it is possible to train a machine learning model to estimate the effects of individual assistance actions for specific practice questions using natural language processing techniques without referring to additional student data~\citep{Zhang2024:Compare}. Such predictions could serve as initial priors and improve the data efficiency of adaptive sampling techniques.

One limitation of reinforcement learning algorithms is that they are confined to selecting from a predefined set of teaching actions established by human domain experts. Notably, our analyses identified questions for which none of the available assistance actions yielded significant improvements in learning outcomes compared to the ``no assistance'' baseline (Table~\ref{tab:anova}). Hence, limitations in the pool of available actions can prevent us from learning effective teaching policies and present an obstacle to effective personalization. To address these limitations, future work will draw inspiration from Never-Ending Learning systems~\citep{Mitchell2018:Never} and equip ITSs with the capability to evaluate teaching actions, optimize teaching policies, and identify targets for teaching action refinements. In this context, Generative AI will enable the automated authoring of new feedback content elements and adaptive scaffolding workflows~\citep{Nguyen2023:Evaluating,Pardos2024:Chatgpt,Letteri2025:Enhancing}. The automated generation and evaluation of feedback at scale will provide a rich environment for research on effective instructional design principles~\citep{Koedinger2013:Instructional}.

\section{Conclusion}
\label{sec:conclusion}

We designed, implemented,  evaluated, and deployed a data-driven system that optimizes feedback policies in a large-scale online tutoring system, supporting students after incorrect responses to practice questions. Analyzing data from one million students, the system learned to provide effective assistance actions (e.g., hints and explanatory text), leading to significant improvements in student learning outcomes. Using offline policy evaluation, we characterized trade-offs in optimizing feedback for different measures of learning outcomes (e.g., reattempt correctness, session completion). Our study further presents the first systematic comparison between the effects of multi-armed bandit (MAB), which optimize feedback for the overall student population, and contextual bandit (CB) policies, which personalize feedback based on individual student attributes. Results suggest that, in the context of the studied tutoring system, while the effects of individual assistance actions may vary among students, these variations may not be substantial enough for personalized feedback policies to significantly outperform well-optimized MAB policies. Building on our findings, future research will explore features of effective feedback content and its interaction with individual student learning outcomes. Today, the deployed system continuously evaluates feedback and optimizes policies guiding future refinements of the online tutoring platform.

\begin{appendices}
\section{System Architecture}
\label{app:system_architecture}

\begin{figure*}[t]
    \centering
    \includegraphics[width=0.92\textwidth]{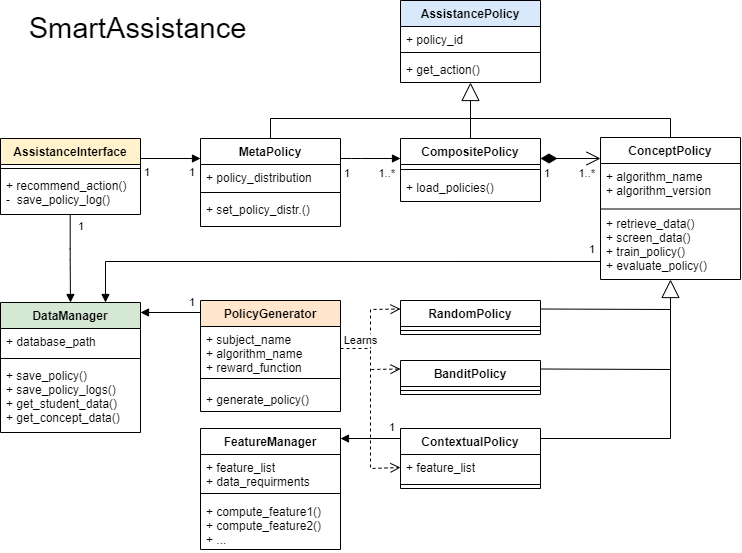}
    \caption{System architecture underlying the SmartAssistance system illustrated as UML class diagram. The system implements a service for the adaptive practice workflow that is called whenever a student requires support. The architecture handles policy training, evaluation, deployment as well as the computation of context features by communicating with a central database.}
    \label{fig:architecture}
\end{figure*}

Here we provide additional details about the software implementation underlying the SmartAssistance system and explain related design considerations. Figure \ref{fig:architecture} illustrates the overall system architecture as UML class diagram. SmartAssistance supports the adaptive practice workflow (described in Section \ref{sec:ck12}) by providing a service that recommends suitable assistance content in response to user-driven and automated assistance queries. SmartAssistance coordinates training, evaluation and deployment of assistance policies and handles related communication with a central database. The modular architecture facilitates the deployment of multi-armed bandit and contextual bandit algorithms and is designed to support reinforcement learning algorithms optimizing sequential instructions in future iterations. All types of assistance policies can be queried via a standardized interface. In the following, we describe the individual system components.

\subsection*{AssistanceInterface}

The \texttt{AsistanceInterface} is an interface class that handles assistance queries from the adaptive practice workflow. To determine an assistance action recommendation, the interface takes as input identifiers describing the lesson and question the learner is currently interacting with. The interface then passes this information to the \texttt{MetaPolicy} object which determines an assistance action identifier. After each assistance action recommendation, the interface writes policy log data relevant for assistance policy training and evaluation to the database.

\subsection*{DataManager}

The \texttt{DataManager} is a helper class that coordinates interactions with a central database hosting student log data, assistance content and trained assistance policies. The \texttt{DataManager} saves assistance policies trained by the \texttt{PolicyGenerator} and records policy log data whenever the \texttt{AssistanceInterace} responds to an assistance query. The \texttt{DataManager} further implements functions to retrieve existing log data for individual learners to provide assistance policies with the necessary information for making personalized assistance recommendations. The \texttt{DataManager} provides the \texttt{PolicyGenerator} with batch access to student log data available for a particular concept to facilitate offline policy optimization and evaluation.

\subsection*{AssistancePolicy}

The \texttt{AssistancePolicy} is an abstract parent class that implements a standardized signature for other assistance policy classes. Each assistance policy object features a unique policy identifier and a get\_action function which takes as input an assistance action query and that in response returns an assistance action identifier.

\subsection*{MetaPolicy}

The \texttt{MetaPolicy} is a class that serves as an abstraction layer that handles interactions between the \texttt{AssistanceInterface} and one or more \texttt{CompositePolicy} objects. Internally, the \texttt{MetaPolicy} features a policy distribution attribute. Whenever the \texttt{MetaPolicy} is queried for an assistance action it randomly selects one of the \texttt{CompositePolicies} objects based on the distribution specified by this attribute. The user-to-policy mapping is implementing via a hash function that operates on an unique practice session identifier which ensures that the policy a student interacts with stays constant during each session. Overall, this enables the orchestration of A/B evaluations by deploying multiple assistance policies in parallel to each other.

\subsection*{CompositePolicy}

The \texttt{CompositePolicy} is a class that manages a set of \texttt{ConceptPolicy} objects. At initialization time the \texttt{CompositePolicy} loads a policy specification file pre-generated from the \texttt{PolicyGenerator} defining one \texttt{ConceptPolicy} for each concept taught by the FlexBooks system into memory. The individual ConceptPolicies can implement different bandit and reinforcement learning algorithms (e.g., multi-armed bandit and contextual bandit algorithms). Whenever the CompositePolicy receives a query for assistance it uses the associated concept identifier to forward the query to the relevant ConceptPolicy to determine and return an assistance action identifier.

\subsection*{ConceptPolicy}

The \texttt{ConceptPolicy} is an abstract class that implements a unified interface inherited by different assistance policy classes (e.g., \texttt{BanditPolicy} and \texttt{ContextualPolicy}). Each \texttt{ConceptPolicy} object features attributes that specify the associated algorithm name and implementation version. The abstract class further defines function signatures for policy training and evaluation as well as related helper functions for retrieval and screening of student log data from the \texttt{DataManager}. Internally, the current implementation of \texttt{ConceptPolicies} features one random or bandit policy for each question tagged to the respective concept. In future work, we want to explore the potential of reinforcement learning algorithms which might be able to leverage synergies between individual assistance actions and questions when making sequential action recommendations.

\subsection*{RandomPolicy}

The \texttt{RandomPolicy} implements a \texttt{ConceptPolicy} class that when queried to recommend assistance for a particular question, uniformly samples an action from the  content pool. The \texttt{RandomPolicy} class was used to perform initial data collection for offline policy evaluations.

\subsection*{BanditPolicy}

The \texttt{BanditPolicy} implements a \texttt{ConceptPolicy} class that when queried to recommend assistance for a particular question, determines an action based on a learned multi-armed bandit policy \citep{Lattimore2020:Bandit}. The multi-armed bandit policies presented in this paper are optimized using offline policy optimization techniques (details in Section \ref{sec:methodology}), but future extensions may consider online learning algorithms that implement adaptive sampling.

\subsection*{ContextualPolicy}

The \texttt{ContextualPolicy} implements a \texttt{ConceptPolicy} class that when queried to recommend assistance for a particular question, determines an action based on a learned contextual bandit policy \citep{Lattimore2020:Bandit}. Each \texttt{ContextualPolicy} object hosts an attribute specifying a list of context features required for selecting assistance. In deployment, the \texttt{ContextualPolicy} retrieves the related student log data from the \texttt{DataManager} and passes it to the FeatureManager to compute the context vector. The contextual bandit policies presented in this paper are optimized using offline policy optimization techniques (details in Section \ref{sec:methodology}), but future extensions may consider online learning algorithms that implement adaptive sampling.

\subsection*{FeatureManager}

The \texttt{FeatureManager} is a helper class that implements a series of functions which take as input student log data to compute different features that provide context for recommending assistance actions. In particular, the \texttt{FeatureManager} plays a central role in the training and evaluation of \texttt{ContextualPolicy} specifications.

\subsection*{PolicyGenerator}

The \texttt{PolicyGenerator} is a helper class that handles the generation of \texttt{CompositePolicy} specification files. In this process, The \texttt{PolicyGenerator} receives a subject and an algorithm name as well as a reward function that guides the policy optimization process (details in Section \ref{sec:methodology}). The \texttt{PolicyGenerator} trains one \texttt{ConceptPolicy} for each concept. For this the \texttt{PolicyGenerator} communicates with the \texttt{DataManager} to retrieve relevant log data and to store resulting policy specification files for later deployment. Policy files are computed at regular intervals as more student log data is observed by the system. Future extensions may consider online learning algorithms for continuous policy refinements during live deployments.

\section{Feature Descriptions}
\label{app:features}

This Appendix serves as reference providing additional implementation details on the individual features used in our contextual bandit experiments \ref{subsec:contextual_methodology}. The aim is to provide guidance to re-implementations efforts. The list of features was informed by prior works on student modeling in intelligent tutoring systems (i.e., \citep{Schmucker2022:Assessing,Prihar2022:Automatic}) and expanded through review of available system usage data and conversations with domain experts.


\begin{table}[!ht]
    \centering
    \caption{Overview of features considered by the contextual assistance policies.}
    \vspace{3mm}
    \begin{tabular}{ll}
        \hline
        Feature & Description \\
        \hline  
        student ability & IRT ability estimate at first attempt. \\
        resp. time & Response time at first attempt. \\
        prev. resp. cor. & Response to prior question correct. \\
        quest. num & Question number in current session. \\
        cor. rate & First attempt correct rate in current session. \\
        \hline
        assigned & Teacher assigned current session. \\
        confidence & Student confidence level ($\{1, 2, 3\}$) at start of session. \\
        weekend & Current session occurs on a weekend. \\
        \hline 
        num sess. total & Number of practice sessions completed overall. \\
        num quest. total & Number of questions answered overall. \\
        num assist. total & Number assistance actions received overall. \\
        \hline
        avg quest. num & Average number of questions answered per session. \\
        avg sess. succ. & Average session success rate ($\geq 10$ correct responses). \\
        \hline
        avg 1st cor. & Average first attempt correct rate. \\
        avg 2nd cor. & Average second attempt correct rate. \\
        \hline
        avg 1st assists. & Average number of assistance actions on first attempts. \\
        avg 2nd assists. & Average number of assistance actions on second attempts. \\
        \hline
        med 1st resp. time & Median response time on first attempts. \\
        med 2nd resp. time & Median response time on second attempts. \\
        med assist. time & Median time spent on assistance actions. \\
        \hline
    \end{tabular}
    \label{tab:feature_overview}
\end{table}

\end{appendices}

\bibliography{bibliography}

\begin{thebibliography}{}
\renewcommand{\doi}[1]{\url{https://doi.org/#1}}
\bibcommenthead

\bibitem [\protect \citeauthoryear {%
Abdelshiheed%
, Hostetter%
, Barnes%
\BCBL {}\ \BBA {} Chi%
}{%
Abdelshiheed%
\ \protect \BOthers {.}}{%
{\protect \APACyear {2023}}%
}]{%
Abdelshiheed2023:Leveraging}
\APACinsertmetastar {%
Abdelshiheed2023:Leveraging}%
\begin{APACrefauthors}%
Abdelshiheed, M.%
, Hostetter, J.W.%
, Barnes, T.%
\BCBL {} Chi, M.%
\end{APACrefauthors}%
\unskip\
\newblock
\APACrefYearMonthDay{2023}{}{}.
\newblock
{\BBOQ}\APACrefatitle {Leveraging deep reinforcement learning for metacognitive
  interventions across intelligent tutoring systems} {Leveraging deep
  reinforcement learning for metacognitive interventions across intelligent
  tutoring systems}.{\BBCQ}
\newblock
 \APACrefbtitle {International Conference on Artificial Intelligence in
  Education} {International conference on artificial intelligence in
  education}\ (\BPGS\ 291--303).
\PrintBackRefs{\CurrentBib}

\bibitem [\protect \citeauthoryear {%
Aleven%
, McLaughlin%
, Glenn%
\BCBL {}\ \BBA {} Koedinger%
}{%
Aleven%
\ \protect \BOthers {.}}{%
{\protect \APACyear {2016}}%
}]{%
Aleven2016:Instruction}
\APACinsertmetastar {%
Aleven2016:Instruction}%
\begin{APACrefauthors}%
Aleven, V.%
, McLaughlin, E.A.%
, Glenn, R.A.%
\BCBL {} Koedinger, K.R.%
\end{APACrefauthors}%
\unskip\
\newblock
\APACrefYearMonthDay{2016}{}{}.
\newblock
{\BBOQ}\APACrefatitle {Instruction based on adaptive learning technologies}
  {Instruction based on adaptive learning technologies}.{\BBCQ}
\newblock
\APACjournalVolNumPages{Handbook of research on learning and
  instruction}{}{}{522--560,}
\newblock

\newblock

\PrintBackRefs{\CurrentBib}

\bibitem [\protect \citeauthoryear {%
Aleven%
, Roll%
, McLaren%
\BCBL {}\ \BBA {} Koedinger%
}{%
Aleven%
\ \protect \BOthers {.}}{%
{\protect \APACyear {2010}}%
}]{%
Aleven2010:Automated}
\APACinsertmetastar {%
Aleven2010:Automated}%
\begin{APACrefauthors}%
Aleven, V.%
, Roll, I.%
, McLaren, B.M.%
\BCBL {} Koedinger, K.R.%
\end{APACrefauthors}%
\unskip\
\newblock
\APACrefYearMonthDay{2010}{}{}.
\newblock
{\BBOQ}\APACrefatitle {Automated, unobtrusive, action-by-action assessment of
  self-regulation during learning with an intelligent tutoring system}
  {Automated, unobtrusive, action-by-action assessment of self-regulation
  during learning with an intelligent tutoring system}.{\BBCQ}
\newblock
\APACjournalVolNumPages{Educational Psychologist}{45}{4}{224--233,}
\newblock

\newblock

\PrintBackRefs{\CurrentBib}

\bibitem [\protect \citeauthoryear {%
Aleven%
, Stahl%
, Schworm%
, Fischer%
\BCBL {}\ \BBA {} Wallace%
}{%
Aleven%
\ \protect \BOthers {.}}{%
{\protect \APACyear {2003}}%
}]{%
Aleven2003:Help}
\APACinsertmetastar {%
Aleven2003:Help}%
\begin{APACrefauthors}%
Aleven, V.%
, Stahl, E.%
, Schworm, S.%
, Fischer, F.%
\BCBL {} Wallace, R.%
\end{APACrefauthors}%
\unskip\
\newblock
\APACrefYearMonthDay{2003}{}{}.
\newblock
{\BBOQ}\APACrefatitle {Help seeking and help design in interactive learning
  environments} {Help seeking and help design in interactive learning
  environments}.{\BBCQ}
\newblock
\APACjournalVolNumPages{Review of educational research}{73}{3}{277--320,}
\newblock

\newblock

\PrintBackRefs{\CurrentBib}

\bibitem [\protect \citeauthoryear {%
Ausin%
, Azizsoltani%
, Barnes%
\BCBL {}\ \BBA {} Chi%
}{%
Ausin%
\ \protect \BOthers {.}}{%
{\protect \APACyear {2019}}%
}]{%
Ausin2019:Leveraging}
\APACinsertmetastar {%
Ausin2019:Leveraging}%
\begin{APACrefauthors}%
Ausin, M.S.%
, Azizsoltani, H.%
, Barnes, T.%
\BCBL {} Chi, M.%
\end{APACrefauthors}%
\unskip\
\newblock
\APACrefYearMonthDay{2019}{}{}.
\newblock
{\BBOQ}\APACrefatitle {Leveraging Deep Reinforcement Learning for Pedagogical
  Policy Induction in an Intelligent Tutoring System} {Leveraging deep
  reinforcement learning for pedagogical policy induction in an intelligent
  tutoring system}.{\BBCQ}
\newblock
 \APACrefbtitle {Proceedings of the 12th International Conference on
  Educational Data Mining} {Proceedings of the 12th international conference on
  educational data mining}\ (\BPGS\ 168--177).
\newblock
\APACaddressPublisher{Montréal, QC, Canada}{EDM}.
\PrintBackRefs{\CurrentBib}

\bibitem [\protect \citeauthoryear {%
Ausin%
, Maniktala%
, Barnes%
\BCBL {}\ \BBA {} Chi%
}{%
Ausin%
\ \protect \BOthers {.}}{%
{\protect \APACyear {2021}}%
}]{%
Ausin2021:Tackling}
\APACinsertmetastar {%
Ausin2021:Tackling}%
\begin{APACrefauthors}%
Ausin, M.S.%
, Maniktala, M.%
, Barnes, T.%
\BCBL {} Chi, M.%
\end{APACrefauthors}%
\unskip\
\newblock
\APACrefYearMonthDay{2021}{}{}.
\newblock
{\BBOQ}\APACrefatitle {Tackling the Credit Assignment Problem in Reinforcement
  Learning-Induced Pedagogical Policies with Neural Networks} {Tackling the
  credit assignment problem in reinforcement learning-induced pedagogical
  policies with neural networks}.{\BBCQ}
\newblock
 \APACrefbtitle {Artificial Intelligence in Education} {Artificial intelligence
  in education}\ (\BPGS\ 356--368).
\newblock
\APACaddressPublisher{Cham}{Springer}.
\PrintBackRefs{\CurrentBib}

\bibitem [\protect \citeauthoryear {%
Baker%
, Corbett%
, Koedinger%
\BCBL {}\ \BBA {} Wagner%
}{%
Baker%
\ \protect \BOthers {.}}{%
{\protect \APACyear {2004}}%
}]{%
Baker2004:Off}
\APACinsertmetastar {%
Baker2004:Off}%
\begin{APACrefauthors}%
Baker, R.S.%
, Corbett, A.T.%
, Koedinger, K.R.%
\BCBL {} Wagner, A.Z.%
\end{APACrefauthors}%
\unskip\
\newblock
\APACrefYearMonthDay{2004}{}{}.
\newblock
{\BBOQ}\APACrefatitle {Off-task behavior in the cognitive tutor classroom: When
  students" game the system"} {Off-task behavior in the cognitive tutor
  classroom: When students" game the system"}.{\BBCQ}
\newblock
 \APACrefbtitle {Proceedings of the SIGCHI conference on Human factors in
  computing systems} {Proceedings of the sigchi conference on human factors in
  computing systems}\ (\BPGS\ 383--390).
\PrintBackRefs{\CurrentBib}

\bibitem [\protect \citeauthoryear {%
Baker%
\ \BBA {} Hawn%
}{%
Baker%
\ \BBA {} Hawn%
}{%
{\protect \APACyear {2022}}%
}]{%
Baker2022:Algorithmic}
\APACinsertmetastar {%
Baker2022:Algorithmic}%
\begin{APACrefauthors}%
Baker, R.S.%
\BCBT {}\ \BBA {} Hawn, A.%
\end{APACrefauthors}%
\unskip\
\newblock
\APACrefYearMonthDay{2022}{}{}.
\newblock
{\BBOQ}\APACrefatitle {Algorithmic bias in education} {Algorithmic bias in
  education}.{\BBCQ}
\newblock
\APACjournalVolNumPages{International Journal of Artificial Intelligence in
  Education}{}{}{1--41,}
\newblock

\newblock

\PrintBackRefs{\CurrentBib}

\bibitem [\protect \citeauthoryear {%
Barnes%
\ \BBA {} Stamper%
}{%
Barnes%
\ \BBA {} Stamper%
}{%
{\protect \APACyear {2008}}%
}]{%
Barnes2008:Toward}
\APACinsertmetastar {%
Barnes2008:Toward}%
\begin{APACrefauthors}%
Barnes, T.%
\BCBT {}\ \BBA {} Stamper, J.%
\end{APACrefauthors}%
\unskip\
\newblock
\APACrefYearMonthDay{2008}{}{}.
\newblock
{\BBOQ}\APACrefatitle {Toward Automatic Hint Generation for Logic Proof
  Tutoring Using Historical Student Data} {Toward automatic hint generation for
  logic proof tutoring using historical student data}.{\BBCQ}
\newblock
 \APACrefbtitle {Intelligent Tutoring Systems} {Intelligent tutoring systems}\
  (\BPGS\ 373--382).
\newblock
\APACaddressPublisher{Berlin, Heidelberg}{Springer Berlin Heidelberg}.
\PrintBackRefs{\CurrentBib}

\bibitem [\protect \citeauthoryear {%
Benjamini%
\ \BBA {} Hochberg%
}{%
Benjamini%
\ \BBA {} Hochberg%
}{%
{\protect \APACyear {1995}}%
}]{%
Benjamini1995:Controlling}
\APACinsertmetastar {%
Benjamini1995:Controlling}%
\begin{APACrefauthors}%
Benjamini, Y.%
\BCBT {}\ \BBA {} Hochberg, Y.%
\end{APACrefauthors}%
\unskip\
\newblock
\APACrefYearMonthDay{1995}{}{}.
\newblock
{\BBOQ}\APACrefatitle {Controlling the false discovery rate: a practical and
  powerful approach to multiple testing} {Controlling the false discovery rate:
  a practical and powerful approach to multiple testing}.{\BBCQ}
\newblock
\APACjournalVolNumPages{Journal of the Royal statistical society: series B
  (Methodological)}{57}{1}{289--300,}
\newblock

\newblock

\PrintBackRefs{\CurrentBib}

\bibitem [\protect \citeauthoryear {%
Bernacki%
, Greene%
\BCBL {}\ \BBA {} Lobczowski%
}{%
Bernacki%
\ \protect \BOthers {.}}{%
{\protect \APACyear {2021}}%
}]{%
Bernacki2021:Systematic}
\APACinsertmetastar {%
Bernacki2021:Systematic}%
\begin{APACrefauthors}%
Bernacki, M.L.%
, Greene, M.J.%
\BCBL {} Lobczowski, N.G.%
\end{APACrefauthors}%
\unskip\
\newblock
\APACrefYearMonthDay{2021}{}{}.
\newblock
{\BBOQ}\APACrefatitle {A systematic review of research on personalized
  learning: Personalized by whom, to what, how, and for what purpose (s)?} {A
  systematic review of research on personalized learning: Personalized by whom,
  to what, how, and for what purpose (s)?}{\BBCQ}
\newblock
\APACjournalVolNumPages{Educational Psychology Review}{33}{4}{1675--1715,}
\newblock

\newblock

\PrintBackRefs{\CurrentBib}

\bibitem [\protect \citeauthoryear {%
Bier%
, Stamper%
, Moore%
, Siegel%
\BCBL {}\ \BBA {} Anbar%
}{%
Bier%
\ \protect \BOthers {.}}{%
{\protect \APACyear {2023}}%
}]{%
Bier2023:Oli}
\APACinsertmetastar {%
Bier2023:Oli}%
\begin{APACrefauthors}%
Bier, N.%
, Stamper, J.%
, Moore, S.%
, Siegel, D.%
\BCBL {} Anbar, A.%
\end{APACrefauthors}%
\unskip\
\newblock
\APACrefYearMonthDay{2023}{}{}.
\newblock
{\BBOQ}\APACrefatitle {OLI Torus: a next-generation, open platform for adaptive
  courseware development, delivery, and research} {Oli torus: a
  next-generation, open platform for adaptive courseware development, delivery,
  and research}.{\BBCQ}
\newblock
 \APACrefbtitle {Companion Proceedings of the 13th International Learning
  Analytics and Knowledge Conference.} {Companion proceedings of the 13th
  international learning analytics and knowledge conference.}
\PrintBackRefs{\CurrentBib}

\bibitem [\protect \citeauthoryear {%
Cameron%
\ \BBA {} Miller%
}{%
Cameron%
\ \BBA {} Miller%
}{%
{\protect \APACyear {2015}}%
}]{%
Cameron2015:Practitioner}
\APACinsertmetastar {%
Cameron2015:Practitioner}%
\begin{APACrefauthors}%
Cameron, A.C.%
\BCBT {}\ \BBA {} Miller, D.L.%
\end{APACrefauthors}%
\unskip\
\newblock
\APACrefYearMonthDay{2015}{}{}.
\newblock
{\BBOQ}\APACrefatitle {A practitioner’s guide to cluster-robust inference} {A
  practitioner’s guide to cluster-robust inference}.{\BBCQ}
\newblock
\APACjournalVolNumPages{Journal of human resources}{50}{2}{317--372,}
\newblock

\newblock

\PrintBackRefs{\CurrentBib}

\bibitem [\protect \citeauthoryear {%
M.~Chi%
, Jordan%
, VanLehn%
\BCBL {}\ \BBA {} Litman%
}{%
M.~Chi%
\ \protect \BOthers {.}}{%
{\protect \APACyear {2009}}%
}]{%
Chi2009:To}
\APACinsertmetastar {%
Chi2009:To}%
\begin{APACrefauthors}%
Chi, M.%
, Jordan, P.W.%
, VanLehn, K.%
\BCBL {} Litman, D.J.%
\end{APACrefauthors}%
\unskip\
\newblock
\APACrefYearMonthDay{2009}{}{}.
\newblock
{\BBOQ}\APACrefatitle {To Elicit Or To Tell: Does It Matter?} {To elicit or to
  tell: Does it matter?}{\BBCQ}
\newblock
 \APACrefbtitle {Proceedings of the 14th International Conference on Artificial
  Intelligence in Education} {Proceedings of the 14th international conference
  on artificial intelligence in education}\ (\BVOL~200, \BPGS\ 197--204).
\newblock
\APACaddressPublisher{Brighton, UK}{{IOS} Press}.
\PrintBackRefs{\CurrentBib}

\bibitem [\protect \citeauthoryear {%
M.~Chi%
, VanLehn%
\BCBL {}\ \BBA {} Litman%
}{%
M.~Chi%
\ \protect \BOthers {.}}{%
{\protect \APACyear {2010}}%
}]{%
Chi2010:Do}
\APACinsertmetastar {%
Chi2010:Do}%
\begin{APACrefauthors}%
Chi, M.%
, VanLehn, K.%
\BCBL {} Litman, D.%
\end{APACrefauthors}%
\unskip\
\newblock
\APACrefYearMonthDay{2010}{}{}.
\newblock
{\BBOQ}\APACrefatitle {Do Micro-Level Tutorial Decisions Matter: Applying
  Reinforcement Learning to Induce Pedagogical Tutorial Tactics} {Do
  micro-level tutorial decisions matter: Applying reinforcement learning to
  induce pedagogical tutorial tactics}.{\BBCQ}
\newblock
 \APACrefbtitle {Intelligent Tutoring Systems} {Intelligent tutoring systems}\
  (\BPGS\ 224--234).
\newblock
\APACaddressPublisher{Berlin, Heidelberg}{Springer Berlin Heidelberg}.
\PrintBackRefs{\CurrentBib}

\bibitem [\protect \citeauthoryear {%
M.T.~Chi%
\ \BBA {} Wylie%
}{%
M.T.~Chi%
\ \BBA {} Wylie%
}{%
{\protect \APACyear {2014}}%
}]{%
Chi2014:ICAP}
\APACinsertmetastar {%
Chi2014:ICAP}%
\begin{APACrefauthors}%
Chi, M.T.%
\BCBT {}\ \BBA {} Wylie, R.%
\end{APACrefauthors}%
\unskip\
\newblock
\APACrefYearMonthDay{2014}{}{}.
\newblock
{\BBOQ}\APACrefatitle {The ICAP framework: Linking cognitive engagement to
  active learning outcomes} {The icap framework: Linking cognitive engagement
  to active learning outcomes}.{\BBCQ}
\newblock
\APACjournalVolNumPages{Educational psychologist}{49}{4}{219--243,}
\newblock

\newblock

\PrintBackRefs{\CurrentBib}

\bibitem [\protect \citeauthoryear {%
De~Ayala%
}{%
De~Ayala%
}{%
{\protect \APACyear {2013}}%
}]{%
Ayala2013:Theory}
\APACinsertmetastar {%
Ayala2013:Theory}%
\begin{APACrefauthors}%
De~Ayala, R.J.%
\end{APACrefauthors}%
\unskip\
\newblock
\APACrefYear{2013}.
\newblock
\APACrefbtitle {The theory and practice of item response theory} {The theory
  and practice of item response theory}.
\newblock
\APACaddressPublisher{New York, NY, USA}{Guilford}.
\PrintBackRefs{\CurrentBib}

\bibitem [\protect \citeauthoryear {%
Doroudi%
, Aleven%
\BCBL {}\ \BBA {} Brunskill%
}{%
Doroudi%
\ \protect \BOthers {.}}{%
{\protect \APACyear {2019}}%
}]{%
Doroudi2019:Where}
\APACinsertmetastar {%
Doroudi2019:Where}%
\begin{APACrefauthors}%
Doroudi, S.%
, Aleven, V.%
\BCBL {} Brunskill, E.%
\end{APACrefauthors}%
\unskip\
\newblock
\APACrefYearMonthDay{2019}{}{}.
\newblock
{\BBOQ}\APACrefatitle {Where’s the reward?} {Where’s the reward?}{\BBCQ}
\newblock
\APACjournalVolNumPages{International Journal of Artificial Intelligence in
  Education}{29}{4}{568--620,}
\newblock

\newblock

\PrintBackRefs{\CurrentBib}

\bibitem [\protect \citeauthoryear {%
Fahid%
\ \protect \BOthers {.}}{%
Fahid%
\ \protect \BOthers {.}}{%
{\protect \APACyear {2021}}%
}]{%
Fahid2021:Adaptively}
\APACinsertmetastar {%
Fahid2021:Adaptively}%
\begin{APACrefauthors}%
Fahid, F.M.%
, Rowe, J.P.%
, Spain, R.D.%
, Goldberg, B.S.%
, Pokorny, R.%
\BCBL {} Lester, J.%
\end{APACrefauthors}%
\unskip\
\newblock
\APACrefYearMonthDay{2021}{}{}.
\newblock
{\BBOQ}\APACrefatitle {Adaptively Scaffolding Cognitive Engagement with Batch
  Constrained Deep Q-Networks} {Adaptively scaffolding cognitive engagement
  with batch constrained deep q-networks}.{\BBCQ}
\newblock
 \APACrefbtitle {Artificial Intelligence in Education} {Artificial intelligence
  in education}\ (\BPGS\ 113--124).
\newblock
\APACaddressPublisher{Cham, CH}{Springer}.
\PrintBackRefs{\CurrentBib}

\bibitem [\protect \citeauthoryear {%
Fancsali%
, Murphy%
\BCBL {}\ \BBA {} Ritter%
}{%
Fancsali%
\ \protect \BOthers {.}}{%
{\protect \APACyear {2022}}%
}]{%
Fancsali2022:Closing}
\APACinsertmetastar {%
Fancsali2022:Closing}%
\begin{APACrefauthors}%
Fancsali, S.%
, Murphy, A.%
\BCBL {} Ritter, S.%
\end{APACrefauthors}%
\unskip\
\newblock
\APACrefYearMonthDay{2022}{July}{}.
\newblock
{\BBOQ}\APACrefatitle {“Closing the Loop” in Educational Data Science with
  an Open Source Architecture for Large-Scale Field Trials} {“closing the
  loop” in educational data science with an open source architecture for
  large-scale field trials}.{\BBCQ}
\newblock
 \APACrefbtitle {Proceedings of the 15th International Conference on
  Educational Data Mining} {Proceedings of the 15th international conference on
  educational data mining}\ (\BPGS\ 834--838).
\newblock
\APACaddressPublisher{Durham, UK}{International Educational Data Mining
  Society}.
\PrintBackRefs{\CurrentBib}

\bibitem [\protect \citeauthoryear {%
Georgila%
\ \protect \BOthers {.}}{%
Georgila%
\ \protect \BOthers {.}}{%
{\protect \APACyear {2019}}%
}]{%
Georgila2019:Using}
\APACinsertmetastar {%
Georgila2019:Using}%
\begin{APACrefauthors}%
Georgila, K.%
, Core, M.G.%
, Nye, B.D.%
, Karumbaiah, S.%
, Auerbach, D.%
\BCBL {} Ram, M.%
\end{APACrefauthors}%
\unskip\
\newblock
\APACrefYearMonthDay{2019}{}{}.
\newblock
{\BBOQ}\APACrefatitle {Using Reinforcement Learning to Optimize the Policies of
  an Intelligent Tutoring System for Interpersonal Skills Training} {Using
  reinforcement learning to optimize the policies of an intelligent tutoring
  system for interpersonal skills training}.{\BBCQ}
\newblock
 \APACrefbtitle {Proceedings of the 18th International Conference on Autonomous
  Agents and MultiAgent Systems} {Proceedings of the 18th international
  conference on autonomous agents and multiagent systems}\ (\BPG~737–745).
\newblock
\APACaddressPublisher{Richland, SC}{AAMAS}.
\PrintBackRefs{\CurrentBib}

\bibitem [\protect \citeauthoryear {%
Heffernan%
\ \BBA {} Heffernan%
}{%
Heffernan%
\ \BBA {} Heffernan%
}{%
{\protect \APACyear {2014}}%
}]{%
Heffernan2014:Assistments}
\APACinsertmetastar {%
Heffernan2014:Assistments}%
\begin{APACrefauthors}%
Heffernan, N.T.%
\BCBT {}\ \BBA {} Heffernan, C.L.%
\end{APACrefauthors}%
\unskip\
\newblock
\APACrefYearMonthDay{2014}{}{}.
\newblock
{\BBOQ}\APACrefatitle {The ASSISTments ecosystem: Building a platform that
  brings scientists and teachers together for minimally invasive research on
  human learning and teaching} {The assistments ecosystem: Building a platform
  that brings scientists and teachers together for minimally invasive research
  on human learning and teaching}.{\BBCQ}
\newblock
\APACjournalVolNumPages{International Journal of Artificial Intelligence in
  Education}{24}{4}{470--497,}
\newblock

\newblock

\PrintBackRefs{\CurrentBib}

\bibitem [\protect \citeauthoryear {%
Ju%
, Zhou%
, Barnes%
\BCBL {}\ \BBA {} Chi%
}{%
Ju%
\ \protect \BOthers {.}}{%
{\protect \APACyear {2020}}%
}]{%
Ju2020:Modeling}
\APACinsertmetastar {%
Ju2020:Modeling}%
\begin{APACrefauthors}%
Ju, S.%
, Zhou, G.%
, Barnes, T.%
\BCBL {} Chi, M.%
\end{APACrefauthors}%
\unskip\
\newblock
\APACrefYearMonthDay{2020}{}{}.
\newblock
{\BBOQ}\APACrefatitle {Pick the Moment: Identifying Critical Pedagogical
  Decisions Using Long-Short Term Rewards.} {Pick the moment: Identifying
  critical pedagogical decisions using long-short term rewards.}{\BBCQ}
\newblock
 \APACrefbtitle {Proceedings of the 13th International Conference on
  Educational Data Mining} {Proceedings of the 13th international conference on
  educational data mining}\ (\BPGS\ 126--136).
\newblock
\APACaddressPublisher{Virtual}{EDM}.
\PrintBackRefs{\CurrentBib}

\bibitem [\protect \citeauthoryear {%
Kizilcec%
\ \BBA {} Lee%
}{%
Kizilcec%
\ \BBA {} Lee%
}{%
{\protect \APACyear {2022}}%
}]{%
Kizilcec2022:Algorithmic}
\APACinsertmetastar {%
Kizilcec2022:Algorithmic}%
\begin{APACrefauthors}%
Kizilcec, R.F.%
\BCBT {}\ \BBA {} Lee, H.%
\end{APACrefauthors}%
\unskip\
\newblock
\APACrefYearMonthDay{2022}{}{}.
\newblock
{\BBOQ}\APACrefatitle {Algorithmic fairness in education} {Algorithmic fairness
  in education}.{\BBCQ}
\newblock
 \APACrefbtitle {The ethics of artificial intelligence in education} {The
  ethics of artificial intelligence in education}\ (\BPGS\ 174--202).
\newblock
\APACaddressPublisher{}{Routledge}.
\PrintBackRefs{\CurrentBib}

\bibitem [\protect \citeauthoryear {%
Koedinger%
\ \BBA {} Aleven%
}{%
Koedinger%
\ \BBA {} Aleven%
}{%
{\protect \APACyear {2007}}%
}]{%
Koedinger2007:Exploring}
\APACinsertmetastar {%
Koedinger2007:Exploring}%
\begin{APACrefauthors}%
Koedinger, K.R.%
\BCBT {}\ \BBA {} Aleven, V.%
\end{APACrefauthors}%
\unskip\
\newblock
\APACrefYearMonthDay{2007}{}{}.
\newblock
{\BBOQ}\APACrefatitle {Exploring the assistance dilemma in experiments with
  cognitive tutors} {Exploring the assistance dilemma in experiments with
  cognitive tutors}.{\BBCQ}
\newblock
\APACjournalVolNumPages{Educational Psychology Review}{19}{3}{239--264,}
\newblock

\newblock

\PrintBackRefs{\CurrentBib}

\bibitem [\protect \citeauthoryear {%
Koedinger%
, Booth%
\BCBL {}\ \BBA {} Klahr%
}{%
Koedinger%
, Booth%
\BCBL {}\ \BBA {} Klahr%
}{%
{\protect \APACyear {2013}}%
}]{%
Koedinger2013:Instructional}
\APACinsertmetastar {%
Koedinger2013:Instructional}%
\begin{APACrefauthors}%
Koedinger, K.R.%
, Booth, J.L.%
\BCBL {} Klahr, D.%
\end{APACrefauthors}%
\unskip\
\newblock
\APACrefYearMonthDay{2013}{}{}.
\newblock
{\BBOQ}\APACrefatitle {Instructional complexity and the science to constrain
  it} {Instructional complexity and the science to constrain it}.{\BBCQ}
\newblock
\APACjournalVolNumPages{Science}{342}{6161}{935--937,}
\newblock

\newblock

\PrintBackRefs{\CurrentBib}

\bibitem [\protect \citeauthoryear {%
Koedinger%
, Brunskill%
, Baker%
, McLaughlin%
\BCBL {}\ \BBA {} Stamper%
}{%
Koedinger%
, Brunskill%
\BCBL {}\ \protect \BOthers {.}}{%
{\protect \APACyear {2013}}%
}]{%
Koedinger2013:New}
\APACinsertmetastar {%
Koedinger2013:New}%
\begin{APACrefauthors}%
Koedinger, K.R.%
, Brunskill, E.%
, Baker, R.S.%
, McLaughlin, E.A.%
\BCBL {} Stamper, J.%
\end{APACrefauthors}%
\unskip\
\newblock
\APACrefYearMonthDay{2013}{}{}.
\newblock
{\BBOQ}\APACrefatitle {New potentials for data-driven intelligent tutoring
  system development and optimization} {New potentials for data-driven
  intelligent tutoring system development and optimization}.{\BBCQ}
\newblock
\APACjournalVolNumPages{AI Magazine}{34}{3}{27--41,}
\newblock

\newblock

\PrintBackRefs{\CurrentBib}

\bibitem [\protect \citeauthoryear {%
Kulik%
\ \BBA {} Fletcher%
}{%
Kulik%
\ \BBA {} Fletcher%
}{%
{\protect \APACyear {2016}}%
}]{%
Kulik2016:Effectiveness}
\APACinsertmetastar {%
Kulik2016:Effectiveness}%
\begin{APACrefauthors}%
Kulik, J.A.%
\BCBT {}\ \BBA {} Fletcher, J.D.%
\end{APACrefauthors}%
\unskip\
\newblock
\APACrefYearMonthDay{2016}{}{}.
\newblock
{\BBOQ}\APACrefatitle {Effectiveness of Intelligent Tutoring Systems: A
  Meta-Analytic Review} {Effectiveness of intelligent tutoring systems: A
  meta-analytic review}.{\BBCQ}
\newblock
\APACjournalVolNumPages{Review of Educational Research}{86}{1}{42-78,}
\newblock

\newblock

\PrintBackRefs{\CurrentBib}

\bibitem [\protect \citeauthoryear {%
Lagoudakis%
\ \BBA {} Parr%
}{%
Lagoudakis%
\ \BBA {} Parr%
}{%
{\protect \APACyear {2003}}%
}]{%
Lagoudakis2003Least}
\APACinsertmetastar {%
Lagoudakis2003Least}%
\begin{APACrefauthors}%
Lagoudakis, M.G.%
\BCBT {}\ \BBA {} Parr, R.%
\end{APACrefauthors}%
\unskip\
\newblock
\APACrefYearMonthDay{2003}{}{}.
\newblock
{\BBOQ}\APACrefatitle {Least-squares policy iteration} {Least-squares policy
  iteration}.{\BBCQ}
\newblock
\APACjournalVolNumPages{The Journal of Machine Learning
  Research}{4}{}{1107--1149,}
\newblock

\newblock

\PrintBackRefs{\CurrentBib}

\bibitem [\protect \citeauthoryear {%
Lattimore%
\ \BBA {} Szepesv{\'a}ri%
}{%
Lattimore%
\ \BBA {} Szepesv{\'a}ri%
}{%
{\protect \APACyear {2020}}%
}]{%
Lattimore2020:Bandit}
\APACinsertmetastar {%
Lattimore2020:Bandit}%
\begin{APACrefauthors}%
Lattimore, T.%
\BCBT {}\ \BBA {} Szepesv{\'a}ri, C.%
\end{APACrefauthors}%
\unskip\
\newblock
\APACrefYear{2020}.
\newblock
\APACrefbtitle {Bandit algorithms} {Bandit algorithms}.
\newblock
\APACaddressPublisher{Cambridge, UK}{Cambridge University Press}.
\PrintBackRefs{\CurrentBib}

\bibitem [\protect \citeauthoryear {%
Leite%
\ \protect \BOthers {.}}{%
Leite%
\ \protect \BOthers {.}}{%
{\protect \APACyear {2022}}%
}]{%
Leite2022:Heterogeneity}
\APACinsertmetastar {%
Leite2022:Heterogeneity}%
\begin{APACrefauthors}%
Leite, W.L.%
, Kuang, H.%
, Shen, Z.%
, Chakraborty, N.%
, Michailidis, G.%
, D'Mello, S.%
\BCBL {} Xing, W.%
\end{APACrefauthors}%
\unskip\
\newblock
\APACrefYearMonthDay{2022}{}{}.
\newblock
{\BBOQ}\APACrefatitle {Heterogeneity of Treatment Effects of a Video
  Recommendation System for Algebra} {Heterogeneity of treatment effects of a
  video recommendation system for algebra}.{\BBCQ}
\newblock
 \APACrefbtitle {Proceedings of the Ninth ACM Conference on Learning @ Scale}
  {Proceedings of the ninth acm conference on learning @ scale}\
  (\BPG~12–23).
\newblock
\APACaddressPublisher{New York, NY, USA}{Association for Computing Machinery}.
\PrintBackRefs{\CurrentBib}

\bibitem [\protect \citeauthoryear {%
Leon%
, Nie%
, Chandak%
\BCBL {}\ \BBA {} Brunskill%
}{%
Leon%
\ \protect \BOthers {.}}{%
{\protect \APACyear {2024}}%
}]{%
Leon2024:Estimating}
\APACinsertmetastar {%
Leon2024:Estimating}%
\begin{APACrefauthors}%
Leon, A.%
, Nie, A.%
, Chandak, Y.%
\BCBL {} Brunskill, E.%
\end{APACrefauthors}%
\unskip\
\newblock
\APACrefYearMonthDay{2024}{}{}.
\newblock
{\BBOQ}\APACrefatitle {Estimating the Causal Treatment Effect of Unproductive
  Persistence} {Estimating the causal treatment effect of unproductive
  persistence}.{\BBCQ}
\newblock
 \APACrefbtitle {Proceedings of the 14th Learning Analytics and Knowledge
  Conference} {Proceedings of the 14th learning analytics and knowledge
  conference}\ (\BPG~843–849).
\newblock
\APACaddressPublisher{New York, NY, USA}{Association for Computing Machinery}.
\PrintBackRefs{\CurrentBib}

\bibitem [\protect \citeauthoryear {%
Letteri%
\ \BBA {} Vittorini%
}{%
Letteri%
\ \BBA {} Vittorini%
}{%
{\protect \APACyear {2025}}%
}]{%
Letteri2025:Enhancing}
\APACinsertmetastar {%
Letteri2025:Enhancing}%
\begin{APACrefauthors}%
Letteri, I.%
\BCBT {}\ \BBA {} Vittorini, P.%
\end{APACrefauthors}%
\unskip\
\newblock
\APACrefYearMonthDay{2025}{}{}.
\newblock
{\BBOQ}\APACrefatitle {Enhancing Student Feedback in Data Science Education:
  Harnessing the Power of AI-Generated Approaches} {Enhancing student feedback
  in data science education: Harnessing the power of ai-generated
  approaches}.{\BBCQ}
\newblock
\APACjournalVolNumPages{International Journal of Artificial Intelligence in
  Education}{}{}{1--24,}
\newblock

\newblock

\PrintBackRefs{\CurrentBib}

\bibitem [\protect \citeauthoryear {%
L.~Li%
, Chu%
, Langford%
\BCBL {}\ \BBA {} Wang%
}{%
L.~Li%
\ \protect \BOthers {.}}{%
{\protect \APACyear {2011}}%
}]{%
Li2011:Unbiased}
\APACinsertmetastar {%
Li2011:Unbiased}%
\begin{APACrefauthors}%
Li, L.%
, Chu, W.%
, Langford, J.%
\BCBL {} Wang, X.%
\end{APACrefauthors}%
\unskip\
\newblock
\APACrefYearMonthDay{2011}{}{}.
\newblock
{\BBOQ}\APACrefatitle {Unbiased Offline Evaluation of Contextual-Bandit-Based
  News Article Recommendation Algorithms} {Unbiased offline evaluation of
  contextual-bandit-based news article recommendation algorithms}.{\BBCQ}
\newblock
 \APACrefbtitle {Proceedings of the Fourth ACM International Conference on Web
  Search and Data Mining} {Proceedings of the fourth acm international
  conference on web search and data mining}\ (\BPG~297–306).
\newblock
\APACaddressPublisher{New York, NY, USA}{Association for Computing Machinery}.
\PrintBackRefs{\CurrentBib}

\bibitem [\protect \citeauthoryear {%
Z.~Li%
\ \protect \BOthers {.}}{%
Z.~Li%
\ \protect \BOthers {.}}{%
{\protect \APACyear {2020}}%
}]{%
Li2020:Getting}
\APACinsertmetastar {%
Li2020:Getting}%
\begin{APACrefauthors}%
Li, Z.%
, Yee, L.%
, Sauerberg, N.%
, Sakson, I.%
, Williams, J.J.%
\BCBL {} Rafferty, A.N.%
\end{APACrefauthors}%
\unskip\
\newblock
\APACrefYearMonthDay{2020}{}{}.
\newblock
{\BBOQ}\APACrefatitle {Getting Too Personal(ized): The Importance of Feature
  Choice in Online Adaptive Algorithms} {Getting too personal(ized): The
  importance of feature choice in online adaptive algorithms}.{\BBCQ}
\newblock
 \APACrefbtitle {Proceedings of the 13th International Conference on
  Educational Data Mining (EDM 2020)} {Proceedings of the 13th international
  conference on educational data mining (edm 2020)}\ (\BPGS\ 159--170).
\newblock
\APACaddressPublisher{}{EDM}.
\PrintBackRefs{\CurrentBib}

\bibitem [\protect \citeauthoryear {%
McLaren%
, Richey%
, Nguyen%
\BCBL {}\ \BBA {} Hou%
}{%
McLaren%
\ \protect \BOthers {.}}{%
{\protect \APACyear {2022}}%
}]{%
Mclaren2022:Instructional}
\APACinsertmetastar {%
Mclaren2022:Instructional}%
\begin{APACrefauthors}%
McLaren, B.M.%
, Richey, J.E.%
, Nguyen, H.%
\BCBL {} Hou, X.%
\end{APACrefauthors}%
\unskip\
\newblock
\APACrefYearMonthDay{2022}{}{}.
\newblock
{\BBOQ}\APACrefatitle {How instructional context can impact learning with
  educational technology: Lessons from a study with a digital learning game}
  {How instructional context can impact learning with educational technology:
  Lessons from a study with a digital learning game}.{\BBCQ}
\newblock
\APACjournalVolNumPages{Computers \& education}{178}{}{1--20,}
\newblock

\newblock

\PrintBackRefs{\CurrentBib}

\bibitem [\protect \citeauthoryear {%
Mills%
, D’Mello%
, Bosch%
\BCBL {}\ \BBA {} Olney%
}{%
Mills%
\ \protect \BOthers {.}}{%
{\protect \APACyear {2015}}%
}]{%
Mills2015:Mind}
\APACinsertmetastar {%
Mills2015:Mind}%
\begin{APACrefauthors}%
Mills, C.%
, D’Mello, S.%
, Bosch, N.%
\BCBL {} Olney, A.M.%
\end{APACrefauthors}%
\unskip\
\newblock
\APACrefYearMonthDay{2015}{}{}.
\newblock
{\BBOQ}\APACrefatitle {Mind wandering during learning with an intelligent
  tutoring system} {Mind wandering during learning with an intelligent tutoring
  system}.{\BBCQ}
\newblock
 \APACrefbtitle {Artificial Intelligence in Education: 17th International
  Conference, AIED 2015, Madrid, Spain, June 22-26, 2015. Proceedings 17}
  {Artificial intelligence in education: 17th international conference, aied
  2015, madrid, spain, june 22-26, 2015. proceedings 17}\ (\BPGS\ 267--276).
\PrintBackRefs{\CurrentBib}

\bibitem [\protect \citeauthoryear {%
Mitchell%
\ \protect \BOthers {.}}{%
Mitchell%
\ \protect \BOthers {.}}{%
{\protect \APACyear {2018}}%
}]{%
Mitchell2018:Never}
\APACinsertmetastar {%
Mitchell2018:Never}%
\begin{APACrefauthors}%
Mitchell, T.%
, Cohen, W.%
, Hruschka, E.%
, Talukdar, P.%
, Yang, B.%
, Betteridge, J.%
\BDBL {}Welling, J.%
\end{APACrefauthors}%
\unskip\
\newblock
\APACrefYearMonthDay{2018}{}{}.
\newblock
{\BBOQ}\APACrefatitle {Never-ending learning} {Never-ending learning}.{\BBCQ}
\newblock
\APACjournalVolNumPages{Communications of the ACM}{61}{5}{103–115,}
\newblock

\newblock

\PrintBackRefs{\CurrentBib}

\bibitem [\protect \citeauthoryear {%
Musabirov%
\ \protect \BOthers {.}}{%
Musabirov%
\ \protect \BOthers {.}}{%
{\protect \APACyear {2024}}%
}]{%
Musabirov2024:Platform}
\APACinsertmetastar {%
Musabirov2024:Platform}%
\begin{APACrefauthors}%
Musabirov, I.%
, Reza, M.%
, Moore, S.%
, Chen, P.%
, Kumar, H.%
, Li, T.%
\BDBL {}Williams, J.J.%
\end{APACrefauthors}%
\unskip\
\newblock
\APACrefYearMonthDay{2024}{}{}.
\newblock
{\BBOQ}\APACrefatitle {Platform-based Adaptive Experimental Research in
  Education: Lessons Learned from Digital Learning Challenge} {Platform-based
  adaptive experimental research in education: Lessons learned from digital
  learning challenge}.{\BBCQ}
\newblock
 \APACrefbtitle {The Fourteenth International Conference on Learning Analytics
  \& Knowledge (LAK24): Learning Analytics in the Age of Artificial
  Intelligence} {The fourteenth international conference on learning analytics
  \& knowledge (lak24): Learning analytics in the age of artificial
  intelligence}\ (\BPGS\ 37--40).
\PrintBackRefs{\CurrentBib}

\bibitem [\protect \citeauthoryear {%
Nagashima%
\ \protect \BOthers {.}}{%
Nagashima%
\ \protect \BOthers {.}}{%
{\protect \APACyear {2022}}%
}]{%
Nagashima2022:Does}
\APACinsertmetastar {%
Nagashima2022:Does}%
\begin{APACrefauthors}%
Nagashima, T.%
, Ling, E.%
, Zheng, B.%
, Bartel, A.N.%
, Silla, E.M.%
, Vest, N.A.%
\BDBL {}Aleven, V.%
\end{APACrefauthors}%
\unskip\
\newblock
\APACrefYearMonthDay{2022}{}{}.
\newblock
{\BBOQ}\APACrefatitle {How does Sustaining and Interleaving Visual Scaffolding
  Help Learners? A Classroom Study with an Intelligent Tutoring System} {How
  does sustaining and interleaving visual scaffolding help learners? a
  classroom study with an intelligent tutoring system}.{\BBCQ}
\newblock
 \APACrefbtitle {Proceedings of the Annual Meeting of the Cognitive Science
  Society} {Proceedings of the annual meeting of the cognitive science
  society}\ (\BVOL~44).
\PrintBackRefs{\CurrentBib}

\bibitem [\protect \citeauthoryear {%
Nathan%
, Koedinger%
\BCBL {}\ \BBA {} Alibali%
}{%
Nathan%
\ \protect \BOthers {.}}{%
{\protect \APACyear {2001}}%
}]{%
Nathan2001:Expert}
\APACinsertmetastar {%
Nathan2001:Expert}%
\begin{APACrefauthors}%
Nathan, M.J.%
, Koedinger, K.R.%
\BCBL {} Alibali, M.W.%
\end{APACrefauthors}%
\unskip\
\newblock
\APACrefYearMonthDay{2001}{}{}.
\newblock
{\BBOQ}\APACrefatitle {Expert blind spot: When content knowledge eclipses
  pedagogical content knowledge} {Expert blind spot: When content knowledge
  eclipses pedagogical content knowledge}.{\BBCQ}
\newblock
 \APACrefbtitle {Proc. of the 3rd Int. Conf. on Cognitive Science} {Proc. of
  the 3rd int. conf. on cognitive science}\ (\BVOL~3, \BPG~644-648).
\newblock
\APACaddressPublisher{Beijing, China}{USTC Press}.
\PrintBackRefs{\CurrentBib}

\bibitem [\protect \citeauthoryear {%
Nguyen%
, Stec%
, Hou%
, Di%
\BCBL {}\ \BBA {} McLaren%
}{%
Nguyen%
\ \protect \BOthers {.}}{%
{\protect \APACyear {2023}}%
}]{%
Nguyen2023:Evaluating}
\APACinsertmetastar {%
Nguyen2023:Evaluating}%
\begin{APACrefauthors}%
Nguyen, H.A.%
, Stec, H.%
, Hou, X.%
, Di, S.%
\BCBL {} McLaren, B.M.%
\end{APACrefauthors}%
\unskip\
\newblock
\APACrefYearMonthDay{2023}{}{}.
\newblock
{\BBOQ}\APACrefatitle {Evaluating chatgpt’s decimal skills and feedback
  generation in a digital learning game} {Evaluating chatgpt’s decimal skills
  and feedback generation in a digital learning game}.{\BBCQ}
\newblock
 \APACrefbtitle {European Conf. on Technology Enhanced Learning} {European
  conf. on technology enhanced learning}\ (\BPGS\ 278--293).
\PrintBackRefs{\CurrentBib}

\bibitem [\protect \citeauthoryear {%
Nie%
, Reuel%
\BCBL {}\ \BBA {} Brunskill%
}{%
Nie%
\ \protect \BOthers {.}}{%
{\protect \APACyear {2023}}%
}]{%
Nie2023:Understanding}
\APACinsertmetastar {%
Nie2023:Understanding}%
\begin{APACrefauthors}%
Nie, A.%
, Reuel, A\BHBI K.%
\BCBL {} Brunskill, E.%
\end{APACrefauthors}%
\unskip\
\newblock
\APACrefYearMonthDay{2023}{}{}.
\newblock
{\BBOQ}\APACrefatitle {Understanding the impact of reinforcement learning
  personalization on subgroups of students in math tutoring} {Understanding the
  impact of reinforcement learning personalization on subgroups of students in
  math tutoring}.{\BBCQ}
\newblock
 \APACrefbtitle {International Conference on Artificial Intelligence in
  Education} {International conference on artificial intelligence in
  education}\ (\BPGS\ 688--694).
\PrintBackRefs{\CurrentBib}

\bibitem [\protect \citeauthoryear {%
Ostrow%
, Heffernan%
\BCBL {}\ \BBA {} Williams%
}{%
Ostrow%
\ \protect \BOthers {.}}{%
{\protect \APACyear {2017}}%
}]{%
Ostrow2017:Tomorrow}
\APACinsertmetastar {%
Ostrow2017:Tomorrow}%
\begin{APACrefauthors}%
Ostrow, K.%
, Heffernan, N.%
\BCBL {} Williams, J.J.%
\end{APACrefauthors}%
\unskip\
\newblock
\APACrefYearMonthDay{2017}{}{}.
\newblock
{\BBOQ}\APACrefatitle {Tomorrow's EdTech Today: Establishing a Learning
  Platform as a Collaborative Research Tool for Sound Science} {Tomorrow's
  edtech today: Establishing a learning platform as a collaborative research
  tool for sound science}.{\BBCQ}
\newblock
\APACjournalVolNumPages{Teachers College Record}{119}{3}{1-36,}
\newblock

\newblock

\PrintBackRefs{\CurrentBib}

\bibitem [\protect \citeauthoryear {%
Pane%
, Griffin%
, McCaffrey%
\BCBL {}\ \BBA {} Karam%
}{%
Pane%
\ \protect \BOthers {.}}{%
{\protect \APACyear {2014}}%
}]{%
Pane2014:Effectiveness}
\APACinsertmetastar {%
Pane2014:Effectiveness}%
\begin{APACrefauthors}%
Pane, J.F.%
, Griffin, B.A.%
, McCaffrey, D.F.%
\BCBL {} Karam, R.%
\end{APACrefauthors}%
\unskip\
\newblock
\APACrefYearMonthDay{2014}{}{}.
\newblock
{\BBOQ}\APACrefatitle {Effectiveness of cognitive tutor algebra I at scale}
  {Effectiveness of cognitive tutor algebra i at scale}.{\BBCQ}
\newblock
\APACjournalVolNumPages{Educational Evaluation and Policy
  Analysis}{36}{2}{127--144,}
\newblock

\newblock

\PrintBackRefs{\CurrentBib}

\bibitem [\protect \citeauthoryear {%
Pardos%
\ \BBA {} Bhandari%
}{%
Pardos%
\ \BBA {} Bhandari%
}{%
{\protect \APACyear {2024}}%
}]{%
Pardos2024:Chatgpt}
\APACinsertmetastar {%
Pardos2024:Chatgpt}%
\begin{APACrefauthors}%
Pardos, Z.A.%
\BCBT {}\ \BBA {} Bhandari, S.%
\end{APACrefauthors}%
\unskip\
\newblock
\APACrefYearMonthDay{2024}{}{}.
\newblock
{\BBOQ}\APACrefatitle {ChatGPT-generated help produces learning gains
  equivalent to human tutor-authored help on mathematics skills}
  {Chatgpt-generated help produces learning gains equivalent to human
  tutor-authored help on mathematics skills}.{\BBCQ}
\newblock
\APACjournalVolNumPages{Plos one}{19}{5}{e0304013,}
\newblock

\newblock

\PrintBackRefs{\CurrentBib}

\bibitem [\protect \citeauthoryear {%
Patikorn%
\ \BBA {} Heffernan%
}{%
Patikorn%
\ \BBA {} Heffernan%
}{%
{\protect \APACyear {2020}}%
}]{%
Patikorn2020:Effectiveness}
\APACinsertmetastar {%
Patikorn2020:Effectiveness}%
\begin{APACrefauthors}%
Patikorn, T.%
\BCBT {}\ \BBA {} Heffernan, N.T.%
\end{APACrefauthors}%
\unskip\
\newblock
\APACrefYearMonthDay{2020}{}{}.
\newblock
{\BBOQ}\APACrefatitle {Effectiveness of Crowd-Sourcing On-Demand Assistance
  from Teachers in Online Learning Platforms} {Effectiveness of crowd-sourcing
  on-demand assistance from teachers in online learning platforms}.{\BBCQ}
\newblock
 \APACrefbtitle {Proceedings of the Seventh ACM Conference on Learning @ Scale}
  {Proceedings of the seventh acm conference on learning @ scale}\
  (\BPG~115–124).
\newblock
\APACaddressPublisher{New York, NY, USA}{Association for Computing Machinery}.
\PrintBackRefs{\CurrentBib}

\bibitem [\protect \citeauthoryear {%
Pham%
, Vanacore%
, Sales%
\BCBL {}\ \BBA {} Gagnon-Bartsch%
}{%
Pham%
\ \protect \BOthers {.}}{%
{\protect \APACyear {2024}}%
}]{%
Pham2024:lool}
\APACinsertmetastar {%
Pham2024:lool}%
\begin{APACrefauthors}%
Pham, D.M.%
, Vanacore, K.P.%
, Sales, A.C.%
\BCBL {} Gagnon-Bartsch, J.A.%
\end{APACrefauthors}%
\unskip\
\newblock
\APACrefYearMonthDay{2024}{}{}.
\newblock
{\BBOQ}\APACrefatitle {LOOL: Towards Personalization with
  Flexible$\backslash$\& Robust Estimation of Heterogeneous Treatment Effects}
  {Lool: Towards personalization with flexible$\backslash$\& robust estimation
  of heterogeneous treatment effects}.{\BBCQ}
\newblock
 \APACrefbtitle {Proceedings of the 17th International Conference on
  Educational Data Mining} {Proceedings of the 17th international conference on
  educational data mining}\ (\BPGS\ 376--384).
\PrintBackRefs{\CurrentBib}

\bibitem [\protect \citeauthoryear {%
Prihar%
, Haim%
, Sales%
\BCBL {}\ \BBA {} Heffernan%
}{%
Prihar%
\ \protect \BOthers {.}}{%
{\protect \APACyear {2022}}%
}]{%
Prihar2022:Automatic}
\APACinsertmetastar {%
Prihar2022:Automatic}%
\begin{APACrefauthors}%
Prihar, E.%
, Haim, A.%
, Sales, A.%
\BCBL {} Heffernan, N.%
\end{APACrefauthors}%
\unskip\
\newblock
\APACrefYearMonthDay{2022}{}{}.
\newblock
{\BBOQ}\APACrefatitle {Automatic Interpretable Personalized Learning}
  {Automatic interpretable personalized learning}.{\BBCQ}
\newblock
 \APACrefbtitle {Proceedings of the Ninth ACM Conference on Learning @ Scale}
  {Proceedings of the ninth acm conference on learning @ scale}\ (\BPG~1–11).
\newblock
\APACaddressPublisher{New York, NY, USA}{ACM}.
\PrintBackRefs{\CurrentBib}

\bibitem [\protect \citeauthoryear {%
Prihar%
, Sales%
\BCBL {}\ \BBA {} Heffernan%
}{%
Prihar%
\ \protect \BOthers {.}}{%
{\protect \APACyear {2023}}%
}]{%
Prihar2023:Bandit}
\APACinsertmetastar {%
Prihar2023:Bandit}%
\begin{APACrefauthors}%
Prihar, E.%
, Sales, A.%
\BCBL {} Heffernan, N.%
\end{APACrefauthors}%
\unskip\
\newblock
\APACrefYearMonthDay{2023}{}{}.
\newblock
{\BBOQ}\APACrefatitle {A Bandit You Can Trust} {A bandit you can trust}.{\BBCQ}
\newblock
 \APACrefbtitle {Proceedings of the 31st ACM Conference on User Modeling,
  Adaptation and Personalization} {Proceedings of the 31st acm conference on
  user modeling, adaptation and personalization}\ (\BPGS\ 106--115).
\PrintBackRefs{\CurrentBib}

\bibitem [\protect \citeauthoryear {%
Rafferty%
, Ying%
\BCBL {}\ \BBA {} Williams%
}{%
Rafferty%
\ \protect \BOthers {.}}{%
{\protect \APACyear {2019}}%
}]{%
Rafferty2019:Statistical}
\APACinsertmetastar {%
Rafferty2019:Statistical}%
\begin{APACrefauthors}%
Rafferty, A.%
, Ying, H.%
\BCBL {} Williams, J.%
\end{APACrefauthors}%
\unskip\
\newblock
\APACrefYearMonthDay{2019}{}{}.
\newblock
{\BBOQ}\APACrefatitle {Statistical Consequences of using Multi-armed Bandits to
  Conduct Adaptive Educational Experiments} {Statistical consequences of using
  multi-armed bandits to conduct adaptive educational experiments}.{\BBCQ}
\newblock
\APACjournalVolNumPages{Journal of Educational Data Mining}{11}{1}{47–79,}
\newblock

\newblock

\PrintBackRefs{\CurrentBib}

\bibitem [\protect \citeauthoryear {%
Reza%
, Kim%
, Bhattacharjee%
, Rafferty%
\BCBL {}\ \BBA {} Williams%
}{%
Reza%
\ \protect \BOthers {.}}{%
{\protect \APACyear {2021}}%
}]{%
Reza2021:MOOclet}
\APACinsertmetastar {%
Reza2021:MOOclet}%
\begin{APACrefauthors}%
Reza, M.%
, Kim, J.%
, Bhattacharjee, A.%
, Rafferty, A.N.%
\BCBL {} Williams, J.J.%
\end{APACrefauthors}%
\unskip\
\newblock
\APACrefYearMonthDay{2021}{}{}.
\newblock
{\BBOQ}\APACrefatitle {The MOOClet Framework: Unifying Experimentation, Dynamic
  Improvement, and Personalization in Online Courses} {The mooclet framework:
  Unifying experimentation, dynamic improvement, and personalization in online
  courses}.{\BBCQ}
\newblock
 \APACrefbtitle {Proceedings of the Eighth ACM Conference on Learning @ Scale}
  {Proceedings of the eighth acm conference on learning @ scale}\
  (\BPG~15–26).
\newblock
\APACaddressPublisher{New York, NY, USA}{Association for Computing Machinery}.
\PrintBackRefs{\CurrentBib}

\bibitem [\protect \citeauthoryear {%
Ritter%
, Joshi%
, Fancsali%
\BCBL {}\ \BBA {} Nixon%
}{%
Ritter%
\ \protect \BOthers {.}}{%
{\protect \APACyear {2013}}%
}]{%
Ritter2013:Predicting}
\APACinsertmetastar {%
Ritter2013:Predicting}%
\begin{APACrefauthors}%
Ritter, S.%
, Joshi, A.%
, Fancsali, S.%
\BCBL {} Nixon, T.%
\end{APACrefauthors}%
\unskip\
\newblock
\APACrefYearMonthDay{2013}{}{}.
\newblock
{\BBOQ}\APACrefatitle {Predicting standardized test scores from cognitive tutor
  interactions} {Predicting standardized test scores from cognitive tutor
  interactions}.{\BBCQ}
\newblock
 \APACrefbtitle {Educational Data Mining 2013.} {Educational data mining 2013.}
\PrintBackRefs{\CurrentBib}

\bibitem [\protect \citeauthoryear {%
Ritter%
, Yudelson%
, Fancsali%
\BCBL {}\ \BBA {} Berman%
}{%
Ritter%
\ \protect \BOthers {.}}{%
{\protect \APACyear {2016}}%
}]{%
Ritter2016:How}
\APACinsertmetastar {%
Ritter2016:How}%
\begin{APACrefauthors}%
Ritter, S.%
, Yudelson, M.%
, Fancsali, S.E.%
\BCBL {} Berman, S.R.%
\end{APACrefauthors}%
\unskip\
\newblock
\APACrefYearMonthDay{2016}{}{}.
\newblock
{\BBOQ}\APACrefatitle {How Mastery Learning Works at Scale} {How mastery
  learning works at scale}.{\BBCQ}
\newblock
 \APACrefbtitle {Proceedings of the Third ACM Conference on Learning @ Scale}
  {Proceedings of the third acm conference on learning @ scale}\
  (\BPG~71–79).
\newblock
\APACaddressPublisher{New York, NY, USA}{ACM}.
\PrintBackRefs{\CurrentBib}

\bibitem [\protect \citeauthoryear {%
Ruan%
\ \protect \BOthers {.}}{%
Ruan%
\ \protect \BOthers {.}}{%
{\protect \APACyear {2024}}%
}]{%
Ruan2024:Reinforcement}
\APACinsertmetastar {%
Ruan2024:Reinforcement}%
\begin{APACrefauthors}%
Ruan, S.%
, Nie, A.%
, Steenbergen, W.%
, He, J.%
, Zhang, J.%
, Guo, M.%
\BDBL {}Brunskill, E.%
\end{APACrefauthors}%
\unskip\
\newblock
\APACrefYearMonthDay{2024}{}{}.
\newblock
{\BBOQ}\APACrefatitle {Reinforcement learning tutor better supported lower
  performers in a math task} {Reinforcement learning tutor better supported
  lower performers in a math task}.{\BBCQ}
\newblock
\APACjournalVolNumPages{Machine Learning}{}{}{1--26,}
\newblock

\newblock

\PrintBackRefs{\CurrentBib}

\bibitem [\protect \citeauthoryear {%
Rubin%
}{%
Rubin%
}{%
{\protect \APACyear {2005}}%
}]{%
Rubin2005:Causal}
\APACinsertmetastar {%
Rubin2005:Causal}%
\begin{APACrefauthors}%
Rubin, D.B.%
\end{APACrefauthors}%
\unskip\
\newblock
\APACrefYearMonthDay{2005}{}{}.
\newblock
{\BBOQ}\APACrefatitle {Causal inference using potential outcomes: Design,
  modeling, decisions} {Causal inference using potential outcomes: Design,
  modeling, decisions}.{\BBCQ}
\newblock
\APACjournalVolNumPages{Journal of the American Statistical
  Association}{100}{469}{322--331,}
\newblock

\newblock

\PrintBackRefs{\CurrentBib}

\bibitem [\protect \citeauthoryear {%
Sales%
, Vanacore%
, Kang%
\BCBL {}\ \BBA {} Whittaker%
}{%
Sales%
\ \protect \BOthers {.}}{%
{\protect \APACyear {2024}}%
}]{%
Sales2024:Problem}
\APACinsertmetastar {%
Sales2024:Problem}%
\begin{APACrefauthors}%
Sales, A.C.%
, Vanacore, K.P.%
, Kang, H\BHBI A.%
\BCBL {} Whittaker, T.A.%
\end{APACrefauthors}%
\unskip\
\newblock
\APACrefYearMonthDay{2024}{July}{}.
\newblock
{\BBOQ}\APACrefatitle {Problem-Solving Behavior and EdTech Effectiveness: A
  Model for Exploratory Causal Analysis} {Problem-solving behavior and edtech
  effectiveness: A model for exploratory causal analysis}.{\BBCQ}
\newblock
 \APACrefbtitle {Proceedings of the 17th International Conference on
  Educational Data Mining} {Proceedings of the 17th international conference on
  educational data mining}\ (\BPGS\ 385--395).
\newblock
\APACaddressPublisher{Atlanta, Georgia, USA}{International Educational Data
  Mining Society}.
\PrintBackRefs{\CurrentBib}

\bibitem [\protect \citeauthoryear {%
Sales%
, Wilks%
\BCBL {}\ \BBA {} Pane%
}{%
Sales%
\ \protect \BOthers {.}}{%
{\protect \APACyear {2016}}%
}]{%
Sales2016:Student}
\APACinsertmetastar {%
Sales2016:Student}%
\begin{APACrefauthors}%
Sales, A.C.%
, Wilks, A.%
\BCBL {} Pane, J.F.%
\end{APACrefauthors}%
\unskip\
\newblock
\APACrefYearMonthDay{2016}{}{}.
\newblock
{\BBOQ}\APACrefatitle {Student Usage Predicts Treatment Effect Heterogeneity in
  the Cognitive Tutor Algebra I Program.} {Student usage predicts treatment
  effect heterogeneity in the cognitive tutor algebra i program.}{\BBCQ}
\newblock
\APACjournalVolNumPages{International Educational Data Mining Society}{}{}{,}
\newblock

\newblock

\PrintBackRefs{\CurrentBib}

\bibitem [\protect \citeauthoryear {%
Schmid%
\ \protect \BOthers {.}}{%
Schmid%
\ \protect \BOthers {.}}{%
{\protect \APACyear {2014}}%
}]{%
Schmid2014:Effects}
\APACinsertmetastar {%
Schmid2014:Effects}%
\begin{APACrefauthors}%
Schmid, R.F.%
, Bernard, R.M.%
, Borokhovski, E.%
, Tamim, R.M.%
, Abrami, P.C.%
, Surkes, M.A.%
\BDBL {}Woods, J.%
\end{APACrefauthors}%
\unskip\
\newblock
\APACrefYearMonthDay{2014}{}{}.
\newblock
{\BBOQ}\APACrefatitle {The effects of technology use in postsecondary
  education: A meta-analysis of classroom applications} {The effects of
  technology use in postsecondary education: A meta-analysis of classroom
  applications}.{\BBCQ}
\newblock
\APACjournalVolNumPages{Computers \& Education}{72}{}{271--291,}
\newblock

\newblock

\PrintBackRefs{\CurrentBib}

\bibitem [\protect \citeauthoryear {%
Schmucker%
, Pachapurkar%
, Bala%
, Shah%
\BCBL {}\ \BBA {} Mitchell%
}{%
Schmucker%
\ \protect \BOthers {.}}{%
{\protect \APACyear {2023}}%
}]{%
Schmucker23:Learning}
\APACinsertmetastar {%
Schmucker23:Learning}%
\begin{APACrefauthors}%
Schmucker, R.%
, Pachapurkar, N.%
, Bala, S.%
, Shah, M.%
\BCBL {} Mitchell, T.%
\end{APACrefauthors}%
\unskip\
\newblock
\APACrefYearMonthDay{2023}{}{}.
\newblock
{\BBOQ}\APACrefatitle {Learning to Give Useful Hints: Assistance Action
  Evaluation and Policy Improvements} {Learning to give useful hints:
  Assistance action evaluation and policy improvements}.{\BBCQ}
\newblock
 \APACrefbtitle {Responsive and Sustainable Educational Futures} {Responsive
  and sustainable educational futures}\ (\BPGS\ 383--398).
\newblock
\APACaddressPublisher{Cham}{Springer Nature Switzerland}.
\PrintBackRefs{\CurrentBib}

\bibitem [\protect \citeauthoryear {%
Schmucker%
, Wang%
, Hu%
\BCBL {}\ \BBA {} Mitchell%
}{%
Schmucker%
\ \protect \BOthers {.}}{%
{\protect \APACyear {2022}}%
}]{%
Schmucker2022:Assessing}
\APACinsertmetastar {%
Schmucker2022:Assessing}%
\begin{APACrefauthors}%
Schmucker, R.%
, Wang, J.%
, Hu, S.%
\BCBL {} Mitchell, T.%
\end{APACrefauthors}%
\unskip\
\newblock
\APACrefYearMonthDay{2022}{}{}.
\newblock
{\BBOQ}\APACrefatitle {Assessing the Knowledge State of Online Students - New
  Data, New Approaches, Improved Accuracy} {Assessing the knowledge state of
  online students - new data, new approaches, improved accuracy}.{\BBCQ}
\newblock
\APACjournalVolNumPages{Journal of Educational Data Mining}{14}{1}{1–45,}
\newblock

\newblock

\PrintBackRefs{\CurrentBib}

\bibitem [\protect \citeauthoryear {%
Schudde%
}{%
Schudde%
}{%
{\protect \APACyear {2018}}%
}]{%
Schudde2018:Heterogeneous}
\APACinsertmetastar {%
Schudde2018:Heterogeneous}%
\begin{APACrefauthors}%
Schudde, L.%
\end{APACrefauthors}%
\unskip\
\newblock
\APACrefYearMonthDay{2018}{}{}.
\newblock
{\BBOQ}\APACrefatitle {Heterogeneous effects in education: The promise and
  challenge of incorporating intersectionality into quantitative methodological
  approaches} {Heterogeneous effects in education: The promise and challenge of
  incorporating intersectionality into quantitative methodological
  approaches}.{\BBCQ}
\newblock
\APACjournalVolNumPages{Review of Research in Education}{42}{1}{72--92,}
\newblock

\newblock

\PrintBackRefs{\CurrentBib}

\bibitem [\protect \citeauthoryear {%
Selent%
, Patikorn%
\BCBL {}\ \BBA {} Heffernan%
}{%
Selent%
\ \protect \BOthers {.}}{%
{\protect \APACyear {2016}}%
}]{%
Selent2016:Assistments}
\APACinsertmetastar {%
Selent2016:Assistments}%
\begin{APACrefauthors}%
Selent, D.%
, Patikorn, T.%
\BCBL {} Heffernan, N.%
\end{APACrefauthors}%
\unskip\
\newblock
\APACrefYearMonthDay{2016}{}{}.
\newblock
{\BBOQ}\APACrefatitle {Assistments dataset from multiple randomized controlled
  experiments} {Assistments dataset from multiple randomized controlled
  experiments}.{\BBCQ}
\newblock
 \APACrefbtitle {Proceedings of the Third (2016) ACM Conference on Learning@
  Scale} {Proceedings of the third (2016) acm conference on learning@ scale}\
  (\BPGS\ 181--184).
\newblock
\APACaddressPublisher{New York, NY, USA}{Association for Computing Machinery}.
\PrintBackRefs{\CurrentBib}

\bibitem [\protect \citeauthoryear {%
Singla%
, Rafferty%
, Radanovic%
\BCBL {}\ \BBA {} Heffernan%
}{%
Singla%
\ \protect \BOthers {.}}{%
{\protect \APACyear {2021}}%
}]{%
Singla2021:Reinforcement}
\APACinsertmetastar {%
Singla2021:Reinforcement}%
\begin{APACrefauthors}%
Singla, A.%
, Rafferty, A.N.%
, Radanovic, G.%
\BCBL {} Heffernan, N.T.%
\end{APACrefauthors}%
\unskip\
\newblock
\APACrefYearMonthDay{2021}{}{}.
\newblock
{\BBOQ}\APACrefatitle {Reinforcement Learning for Education: Opportunities and
  Challenges} {Reinforcement learning for education: Opportunities and
  challenges}.{\BBCQ}
\newblock
\APACjournalVolNumPages{CoRR}{abs/2107.08828}{}{,}
\newblock
\begin{APACrefURL} {https://arxiv.org/abs/2107.08828} \end{APACrefURL}
\newblock
{\href{https://arxiv.org/abs/2107.08828}{{2107.08828}}}
\newblock

\PrintBackRefs{\CurrentBib}

\bibitem [\protect \citeauthoryear {%
Spain%
\ \protect \BOthers {.}}{%
Spain%
\ \protect \BOthers {.}}{%
{\protect \APACyear {2019}}%
}]{%
Spain2019:Enhancing}
\APACinsertmetastar {%
Spain2019:Enhancing}%
\begin{APACrefauthors}%
Spain, R.%
, Rowe, J.%
, Goldberg, B.%
, Pokorny, R.%
, Lester, J.%
\BCBL {} Rockville, M.%
\end{APACrefauthors}%
\unskip\
\newblock
\APACrefYearMonthDay{2019}{}{}.
\newblock
{\BBOQ}\APACrefatitle {Enhancing learning outcomes through adaptive remediation
  with GIFT} {Enhancing learning outcomes through adaptive remediation with
  gift}.{\BBCQ}
\newblock
 \APACrefbtitle {Proceedings of the 2019 interservice/industry training
  simulation and education conference (I/ITSEC)} {Proceedings of the 2019
  interservice/industry training simulation and education conference
  (i/itsec)}\ (\BPGS\ 1--11).
\newblock
\APACaddressPublisher{Orlando, Florida}{I/ITSEC}.
\PrintBackRefs{\CurrentBib}

\bibitem [\protect \citeauthoryear {%
Spain%
\ \protect \BOthers {.}}{%
Spain%
\ \protect \BOthers {.}}{%
{\protect \APACyear {2021}}%
}]{%
Spain2021:Decision}
\APACinsertmetastar {%
Spain2021:Decision}%
\begin{APACrefauthors}%
Spain, R.%
, Rowe, J.%
, Smith, A.%
, Goldberg, B.%
, Pokorny, R.%
, Mott, B.%
\BCBL {} Lester, J.%
\end{APACrefauthors}%
\unskip\
\newblock
\APACrefYearMonthDay{2021}{}{}.
\newblock
{\BBOQ}\APACrefatitle {A reinforcement learning approach to adaptive
  remediation in online training} {A reinforcement learning approach to
  adaptive remediation in online training}.{\BBCQ}
\newblock
\APACjournalVolNumPages{The Journal of Defense Modeling and
  Simulation}{2}{19}{173–193,}
\newblock

\newblock

\PrintBackRefs{\CurrentBib}

\bibitem [\protect \citeauthoryear {%
Tibshirani%
, Athey%
, Sverdrup%
\BCBL {}\ \BBA {} Wager%
}{%
Tibshirani%
\ \protect \BOthers {.}}{%
{\protect \APACyear {2021}}%
}]{%
Tibshirani2021:GRF}
\APACinsertmetastar {%
Tibshirani2021:GRF}%
\begin{APACrefauthors}%
Tibshirani, J.%
, Athey, S.%
, Sverdrup, E.%
\BCBL {} Wager, S.%
\end{APACrefauthors}%
\unskip\
\newblock
\APACrefYearMonthDay{2021}{}{}.
\newblock
\APACrefbtitle {grf: Generalized Random Forests. R package version 2.0. 2.}
  {grf: Generalized random forests. r package version 2.0. 2.}
\PrintBackRefs{\CurrentBib}

\bibitem [\protect \citeauthoryear {%
K.~Vanacore%
, Sales%
, Liu%
\BCBL {}\ \BBA {} Ottmar%
}{%
K.~Vanacore%
\ \protect \BOthers {.}}{%
{\protect \APACyear {2023}}%
}]{%
Vanacore2023:Benefit}
\APACinsertmetastar {%
Vanacore2023:Benefit}%
\begin{APACrefauthors}%
Vanacore, K.%
, Sales, A.%
, Liu, A.%
\BCBL {} Ottmar, E.%
\end{APACrefauthors}%
\unskip\
\newblock
\APACrefYearMonthDay{2023}{}{}.
\newblock
{\BBOQ}\APACrefatitle {Benefit of gamification for persistent learners:
  Propensity to replay problems moderates algebra-game effectiveness} {Benefit
  of gamification for persistent learners: Propensity to replay problems
  moderates algebra-game effectiveness}.{\BBCQ}
\newblock
 \APACrefbtitle {Proceedings of the Tenth ACM Conference on Learning@ Scale}
  {Proceedings of the tenth acm conference on learning@ scale}\ (\BPGS\
  164--173).
\PrintBackRefs{\CurrentBib}

\bibitem [\protect \citeauthoryear {%
K.P.~Vanacore%
, Gurung%
, Sales%
\BCBL {}\ \BBA {} Heffernan%
}{%
K.P.~Vanacore%
\ \protect \BOthers {.}}{%
{\protect \APACyear {2024}}%
}]{%
Vanacore2024:Effect}
\APACinsertmetastar {%
Vanacore2024:Effect}%
\begin{APACrefauthors}%
Vanacore, K.P.%
, Gurung, A.%
, Sales, A.%
\BCBL {} Heffernan, N.%
\end{APACrefauthors}%
\unskip\
\newblock
\APACrefYearMonthDay{2024}{}{}.
\newblock
{\BBOQ}\APACrefatitle {Effect of Gamification on Gamers: Evaluating
  Interventions for Students Who Game the System: Evaluating Interventions for
  Students Who Gaming the System} {Effect of gamification on gamers: Evaluating
  interventions for students who game the system: Evaluating interventions for
  students who gaming the system}.{\BBCQ}
\newblock
\APACjournalVolNumPages{Journal of Educational Data Mining}{16}{1}{112--140,}
\newblock

\newblock

\PrintBackRefs{\CurrentBib}

\bibitem [\protect \citeauthoryear {%
VanLehn%
}{%
VanLehn%
}{%
{\protect \APACyear {2006}}%
}]{%
Vanlehn2006:Behavior}
\APACinsertmetastar {%
Vanlehn2006:Behavior}%
\begin{APACrefauthors}%
VanLehn, K.%
\end{APACrefauthors}%
\unskip\
\newblock
\APACrefYearMonthDay{2006}{}{}.
\newblock
{\BBOQ}\APACrefatitle {The behavior of tutoring systems} {The behavior of
  tutoring systems}.{\BBCQ}
\newblock
\APACjournalVolNumPages{International journal of artificial intelligence in
  education}{16}{3}{227--265,}
\newblock

\newblock

\PrintBackRefs{\CurrentBib}

\bibitem [\protect \citeauthoryear {%
VanLehn%
}{%
VanLehn%
}{%
{\protect \APACyear {2011}}%
}]{%
Vanlehn2011:Relative}
\APACinsertmetastar {%
Vanlehn2011:Relative}%
\begin{APACrefauthors}%
VanLehn, K.%
\end{APACrefauthors}%
\unskip\
\newblock
\APACrefYearMonthDay{2011}{}{}.
\newblock
{\BBOQ}\APACrefatitle {The Relative Effectiveness of Human Tutoring,
  Intelligent Tutoring Systems, and Other Tutoring Systems} {The relative
  effectiveness of human tutoring, intelligent tutoring systems, and other
  tutoring systems}.{\BBCQ}
\newblock
\APACjournalVolNumPages{Educational Psychologist}{46}{4}{197-221,}
\newblock

\newblock

\PrintBackRefs{\CurrentBib}

\bibitem [\protect \citeauthoryear {%
VanLehn%
\ \protect \BOthers {.}}{%
VanLehn%
\ \protect \BOthers {.}}{%
{\protect \APACyear {2007}}%
}]{%
Vanlehn2007:Tutorial}
\APACinsertmetastar {%
Vanlehn2007:Tutorial}%
\begin{APACrefauthors}%
VanLehn, K.%
, Graesser, A.C.%
, Jackson, G.T.%
, Jordan, P.%
, Olney, A.%
\BCBL {} Ros{\'e}, C.P.%
\end{APACrefauthors}%
\unskip\
\newblock
\APACrefYearMonthDay{2007}{}{}.
\newblock
{\BBOQ}\APACrefatitle {When are tutorial dialogues more effective than
  reading?} {When are tutorial dialogues more effective than reading?}{\BBCQ}
\newblock
\APACjournalVolNumPages{Cognitive science}{31}{1}{3--62,}
\newblock

\newblock

\PrintBackRefs{\CurrentBib}

\bibitem [\protect \citeauthoryear {%
Wager%
\ \BBA {} Athey%
}{%
Wager%
\ \BBA {} Athey%
}{%
{\protect \APACyear {2018}}%
}]{%
Wager2018:Estimation}
\APACinsertmetastar {%
Wager2018:Estimation}%
\begin{APACrefauthors}%
Wager, S.%
\BCBT {}\ \BBA {} Athey, S.%
\end{APACrefauthors}%
\unskip\
\newblock
\APACrefYearMonthDay{2018}{}{}.
\newblock
{\BBOQ}\APACrefatitle {Estimation and inference of heterogeneous treatment
  effects using random forests} {Estimation and inference of heterogeneous
  treatment effects using random forests}.{\BBCQ}
\newblock
\APACjournalVolNumPages{Journal of the American Statistical
  Association}{113}{523}{1228--1242,}
\newblock

\newblock

\PrintBackRefs{\CurrentBib}

\bibitem [\protect \citeauthoryear {%
Williams%
\ \protect \BOthers {.}}{%
Williams%
\ \protect \BOthers {.}}{%
{\protect \APACyear {2016}}%
}]{%
Williams2016:AXIS}
\APACinsertmetastar {%
Williams2016:AXIS}%
\begin{APACrefauthors}%
Williams, J.J.%
, Kim, J.%
, Rafferty, A.%
, Maldonado, S.%
, Gajos, K.Z.%
, Lasecki, W.S.%
\BCBL {} Heffernan, N.%
\end{APACrefauthors}%
\unskip\
\newblock
\APACrefYearMonthDay{2016}{}{}.
\newblock
{\BBOQ}\APACrefatitle {AXIS: Generating Explanations at Scale with
  Learnersourcing and Machine Learning} {Axis: Generating explanations at scale
  with learnersourcing and machine learning}.{\BBCQ}
\newblock
 \APACrefbtitle {Proceedings of the Third (2016) ACM Conference on Learning @
  Scale} {Proceedings of the third (2016) acm conference on learning @ scale}\
  (\BPG~379–388).
\newblock
\APACaddressPublisher{New York, NY, USA}{Association for Computing Machinery}.
\PrintBackRefs{\CurrentBib}

\bibitem [\protect \citeauthoryear {%
Williams%
\ \protect \BOthers {.}}{%
Williams%
\ \protect \BOthers {.}}{%
{\protect \APACyear {2018}}%
}]{%
Williams2018:Enhancing}
\APACinsertmetastar {%
Williams2018:Enhancing}%
\begin{APACrefauthors}%
Williams, J.J.%
, Rafferty, A.N.%
, Tingley, D.%
, Ang, A.%
, Lasecki, W.S.%
\BCBL {} Kim, J.%
\end{APACrefauthors}%
\unskip\
\newblock
\APACrefYearMonthDay{2018}{}{}.
\newblock
{\BBOQ}\APACrefatitle {Enhancing Online Problems Through Instructor-Centered
  Tools for Randomized Experiments} {Enhancing online problems through
  instructor-centered tools for randomized experiments}.{\BBCQ}
\newblock
 \APACrefbtitle {Proceedings of the 2018 CHI Conference on Human Factors in
  Computing Systems} {Proceedings of the 2018 chi conference on human factors
  in computing systems}\ (\BPG~1–12).
\newblock
\APACaddressPublisher{New York, NY, USA}{Association for Computing Machinery}.
\PrintBackRefs{\CurrentBib}

\bibitem [\protect \citeauthoryear {%
Yadlowsky%
, Fleming%
, Shah%
, Brunskill%
\BCBL {}\ \BBA {} Wager%
}{%
Yadlowsky%
\ \protect \BOthers {.}}{%
{\protect \APACyear {2024}}%
}]{%
Yadlowsky2024:Evaluating}
\APACinsertmetastar {%
Yadlowsky2024:Evaluating}%
\begin{APACrefauthors}%
Yadlowsky, S.%
, Fleming, S.%
, Shah, N.%
, Brunskill, E.%
\BCBL {} Wager, S.%
\end{APACrefauthors}%
\unskip\
\newblock
\APACrefYearMonthDay{2024}{}{}.
\newblock
{\BBOQ}\APACrefatitle {Evaluating Treatment Prioritization Rules via
  Rank-Weighted Average Treatment Effects} {Evaluating treatment prioritization
  rules via rank-weighted average treatment effects}.{\BBCQ}
\newblock
\APACjournalVolNumPages{Journal of the American Statistical
  Association}{0}{ja}{1--25,}
\newblock

\newblock

\PrintBackRefs{\CurrentBib}

\bibitem [\protect \citeauthoryear {%
Zhang%
, Kumar%
, Schmucker%
, Azaria%
\BCBL {}\ \BBA {} Mitchell%
}{%
Zhang%
\ \protect \BOthers {.}}{%
{\protect \APACyear {2024}}%
}]{%
Zhang2024:Compare}
\APACinsertmetastar {%
Zhang2024:Compare}%
\begin{APACrefauthors}%
Zhang, T.%
, Kumar, H.A.%
, Schmucker, R.%
, Azaria, A.%
\BCBL {} Mitchell, T.%
\end{APACrefauthors}%
\unskip\
\newblock
\APACrefYearMonthDay{2024}{}{}.
\newblock
{\BBOQ}\APACrefatitle {Learning to Compare Hints: Combining Insights from
  Student Logs and Large Language Models} {Learning to compare hints: Combining
  insights from student logs and large language models}.{\BBCQ}
\newblock
 \APACrefbtitle {Proceedings of the 2024 AAAI Conference on Artificial
  Intelligence} {Proceedings of the 2024 aaai conference on artificial
  intelligence}\ (\BVOL~257, \BPGS\ 162--169).
\newblock
\APACaddressPublisher{}{PMLR}.
\PrintBackRefs{\CurrentBib}

\end{thebibliography}


\backmatter

\bmhead{Acknowledgments}

This research was supported in part by the Air Force Office of Scientific Research (AFOSR) under awards FA95502010118 and FA95501710218.

\end{document}